%% file: acl2023.tex
\pdfoutput=1

\documentclass[11pt]{article}

\usepackage[]{ACL2023}

\usepackage{times}
\usepackage{latexsym}

\usepackage[T1]{fontenc}

\usepackage[utf8]{inputenc}

\usepackage{microtype}
\usepackage{times}
\usepackage{latexsym}
\usepackage{amssymb}
\usepackage{amsfonts}
\usepackage{amsmath} 
\usepackage{amsthm} 
\usepackage{booktabs}
\usepackage{enumerate}
\usepackage{enumitem}
\usepackage{graphicx}
\usepackage{subfigure}
\usepackage{xspace}
\usepackage{float}
\usepackage{bbm}
\usepackage{bm}
\usepackage{multirow}
\usepackage{booktabs}
\usepackage{color}
\usepackage{framed}
\usepackage{stfloats}
\usepackage{iitem}
\usepackage{makecell}
\usepackage{float}
\usepackage{placeins}
\usepackage{afterpage}
\usepackage{bbding}
\usepackage[normalem]{ulem}
\usepackage{soul}
\usepackage{xcolor}
\usepackage{color}
\usepackage{colortbl}
\usepackage{algorithm}
\usepackage{algorithmic}
\usepackage{array}
\usepackage{hyperref}
\usepackage{listings}
\usepackage{lipsum,mwe,cuted}
\usepackage{ragged2e}
\usepackage{framed}
\usepackage{mdframed}
\usepackage{pifont}
\usepackage[most]{tcolorbox}
\usepackage{colortbl}

\newcommand{\paratitle}[1]{\vspace{1.5ex}\noindent\textbf{#1}}
\newcommand{\ie}{\emph{i.e.,}\xspace}

\newcommand{\eg}{\emph{e.g.,}\xspace}

\newcommand{\ignore}[1]{}

\usepackage{makecell}
\usepackage{cite}
\usepackage{xcolor}
\definecolor{tOrange}{RGB}{210,105,30}
\definecolor{tBlue}{rgb}{0.39,0.58,0.93}
\definecolor{tPink}{RGB}{255,20,147}
\definecolor{tGreen}{RGB}{50,205,50}
\definecolor{tGold}{RGB}{255,215,150}

\title{\emph{The Dawn After the Dark:} An Empirical Study on Factuality Hallucination in Large Language Models}

\author{
	Junyi Li\textsuperscript{\rm{1,3}}, 
	Jie Chen\textsuperscript{\rm{1}},
        Ruiyang Ren\textsuperscript{\rm{1}}, 
        Xiaoxue Cheng\textsuperscript{\rm{1}\ },
        Wayne Xin Zhao\textsuperscript{\rm{1}\ }, \\
	\textbf{Jian-Yun Nie}\textsuperscript{\rm{3}} {\rm and} 
	\textbf{Ji-Rong Wen}\textsuperscript{\rm{1,2}} \\
	\textsuperscript{1}Gaoling School of Artificial Intelligence, Renmin University of China \\
	\textsuperscript{2}School of Information, Renmin University of China \\
	\textsuperscript{3}DIRO, Universit\'{e} de Montr\'{e}al \\ 
	\texttt{\{lijunyi, ptyzchenjie\}@ruc.edu.cn} \quad
	\texttt{batmanfly@gmail.com} \\
}

\begin{document}
\maketitle
\sloppy
\begin{abstract}
In the era of large language models~(LLMs), hallucination (\ie the tendency to generate factually incorrect content) poses great  challenge to trustworthy and reliable deployment of LLMs in real-world applications. To tackle the LLM hallucination, three key questions should be well studied: how to detect hallucinations (\emph{detection}), why do LLMs hallucinate (\emph{source}),  and what can be done to mitigate them  (\emph{mitigation}). 
To address these challenges, this work presents a systematic empirical study on LLM hallucination, focused on the the three aspects of hallucination detection, source and mitigation. 
Specially, we construct a new hallucination benchmark \emph{HaluEval 2.0}, and designs a simple yet effective detection method for LLM hallucination. Furthermore, we zoom into the different training or utilization stages of LLMs and extensively analyze the potential factors that lead to the LLM hallucination. Finally, we implement and examine a series of widely used techniques to mitigate the hallucinations in LLMs. 
Our work has led to several important findings to understand the hallucination origin and mitigate the hallucinations in LLMs. 
Our code and data can be accessed at \url{https://github.com/RUCAIBox/HaluEval-2.0}.
\end{abstract}

\input{sec/intro}
\input{sec/Background}

\input{sec/settings.tex}

\input{sec/detection.tex}
\input{sec/source.tex}
\input{sec/mitigation.tex}

\input{sec/related.tex}
\input{sec/conclusions.tex}

\section*{Contributions}

We list the contributions of student authors as follows:

$\bullet$~Junyi Li: Project leader; Full paper writing and revising; Decide the taxonomy of hallucinations for LLMs; Hallucination detection approach and evaluation metric design; Human labelers hiring and scheduling; Experimental design for each section; Figure and table design.

$\bullet$~Jie Chen: Cleaning and construction of the dataset for factual hallucination evaluation; Code implementation of the GPT-4 based hallucination evaluation framework including Fact Extraction, Fact Judgement, Test of Reliability and Evaluation; Code implementation of Effective Decoding (Greedy-nucleus sampling, Factual-nucleus sampling); Implementation of generative code for data used in RLHF; Paper writing of related work; Run experiments of Fact Extraction, Fact Judgement and Evaluation.

$\bullet$~Ruiyang Ren: Participating in the design of the outline, benchmark, and evaluation metrics; Designing and implementing experiments of retrieval augmentation and frequency of pre-training knowledge; Paper writing of retrieval augmentation part.

$\bullet$~Xiaoxue Cheng: Designing and implementation of experiments for Prompt Design, Self-Reflexion, Prompt Improvement, and part of Supervised Fine-Tuning; Paper writing of self-reflexion part.


\bibliography{custom}
\bibliographystyle{acl_natbib}

\appendix



\end{document}

%% file: sec/intro.tex
\section{Introduction}

Large language models (LLMs)~\citep{zhao2023survey} have shown remarkable potential in a wide range of natural language processing (NLP) applications~\citep{brown2020language,Ouyang-arxiv-2022-Training,OpenAI-OpenAI-2023-GPT-4}. However, despite the significant improvement in model capacity, a persistent challenge lies in their tendency to \emph{hallucinate}, \ie generate the content {that looks plausible but is factually incorrect}~\citep{Huang-arxiv-2023-A,ji2023survey,Zhang-CoRR-2023-Siren}. This issue severely restricts the deployment of LLMs in real-world applications (\eg clinical diagnoses), where the reliable generation of trustworthy text is of utmost importance. 

In the era of LLMs, there has been a significant surge of research interest in hallucinations~\citep{yao2023llm,das2023diving,dhuliawala2023chain,varshney2023stitch,manakul2023selfcheckgpt,shi2023trusting}. These studies are mainly centered around three interleaved questions, \ie \emph{why do LLMs hallucinate, how to detect hallucinations, and what can be done to mitigate them?} 
The three key questions pose great challenges to the research community, while existing empirical work mostly focuses on analyzing or addressing individual challenges, still lacking a systematic and in-depth experimental study on LLM hallucinations. 
To tackle these challenges, a more comprehensive analysis is needed to thoroughly research the aforementioned three questions. 

For deciphering the mystery of hallucination in LLMs, we aim to conduct a comprehensive and systematic empirical study on hallucination \emph{detection}, \emph{source}, and \emph{mitigation}. In particular, we mainly focus on studying \emph{factuality hallucination}, which has become one of the primary erroneous sources for LLMs~\citep{Huang-arxiv-2023-A,Zhang-CoRR-2023-Siren}. 
To carry out our research, we zoom into the different stages to train and use LLMs, including pre-training, supervised fine-tuning~(SFT), reinforcement learning from human feedback~(RLHF), and inference, and thus can conduct more in-depth analysis of potential impact of each stage on model hallucination. This analysis approach is quite different from prior work, where they mostly  study the impact of individual stages or strategies to attribute or mitigate the hallucinations.    


To conduct our empirical study, we first 
extend previous work~\citep{Li-arxiv-2023-HaluEval} and construct a new benchmark \textbf{HaluEval 2.0} for evaluating the factuality hallucination in LLMs. Our benchmark contains 8,770 questions from five domains including biomedicine, finance, science, education, and open domain. 
Based on this dataset, we propose a simple yet effective framework for detecting factual errors in LLM responses. Following prior work~\citep{Chern-2023-CoRR-FacTool,Dhuliawala-2023-CoRR-Chain}, our approach decomposes the problem of hallucination detection into two simple sub-problems, \ie extract factual statements from LLM's responses and then determine whether these statements contain hallucinations. Based on the above detection framework, we perform a series of experiments to explore the hallucination sources from four aspects, including pre-training, supervised fine-tuning, prompt design, and inference methods. For each aspect, we extensively examine the effect of possible factors on LLM hallucinations. 
Furthermore, we delve into hallucination mitigation for LLMs through a series of widely used techniques, including reinforcement learning from human feedback (RLHF), retrieval augmentation, self-reflexion, advanced decoding, and prompt improvement. 

It is generally believed to be challenging to deeply understand and fully mitigate LLM hallucinations. By providing more comprehensive empirical analysis, our work aims to further push forward the research on LLM hallucination. The major findings of our empirical study are summarized below: 

$\bullet$ \textbf{Pre-training:} Pre-training on more tokens has a marginal effect on reducing LLM hallucinations, while incorporating specialized data (\eg scientific text) into pre-training can extremely alleviate hallucinations in specific domains. The frequency of pre-training knowledge significantly influences the source of hallucinations, \ie lower frequency, more hallucinations.

$\bullet$ \textbf{Fine-tuning:} Supervised fine-tuning LLMs with improved instructions can be useful to alleviate the hallucinations. Balancing the complexity of instructions is beneficial to reduce the hallucination, while using overly complex instructions results in a higher level of hallucinations. RLHF is an effective method to mitigate LLM halluciantions, while this effect relies on domains.

$\bullet$ \textbf{Inference:} Diversity-oriented decoding methods induce more hallucinations in professional domains, while greedy search exacerbates in open-ended domains. The token-by-token generation manner makes LLMs over-commit to the previous generation mistakes and leads to hallucinations. Quantization, despite accelerating the inference, largely elicits the LLM hallucinations.

$\bullet$ \textbf{Prompt Design:} Incorporating more details into the task description and leveraging in-context learning can decrease the level of hallucinations. Rewriting question or placing task description after the question induces more hallucinations. LLMs show a lower tendency to produce hallucinations on questions that are easier to read and use more formal and specific language.

\begin{table*}[t!]
\small
    \centering
    \begin{tabular}{lp{3.9cm}p{8.5cm}}
    \toprule
       \rowcolor[gray]{.92} \textbf{Type} & \textbf{Question} & \multicolumn{1}{c}{\textbf{LLM Response}} \\
    \midrule
        \multirow{6}{*}{\begin{tabular}[x]{@{}l@{}} \textbf{Entity-error}\\\textbf{Hallucination} \end{tabular}} & When was the Nokia 3510 phone announced? & The Nokia 3510 mobile phone was announced in \textcolor{red}{October 2002}. It was a popular mobile phone atthe time, known for its compact design and features such as a color display and customizablecovers.  \\
    \cmidrule{2-3}
        & Which merican voice actor wrote Iron Man: Rise of Technovore? & The American voice actor involved in English dubs of Japanese anime as well as cartoons who wrote ``Iron Man: Rise of Technovore'' is \textcolor{red}{Brendan Burch}. \\
    \midrule
        \multirow{5.5}{*}{\begin{tabular}[x]{@{}l@{}} \textbf{Relation-error}\\\textbf{Hallucination} \end{tabular}} & What are the chemical elements that make up the human body? & There are dozens of chemical elements that make up the human body, including hydrogen, oxygen, calcium, iron, and potassium. Among these chemical elements, \textcolor{red}{calcium is more than oxygen}. \\
    \cmidrule{2-3}
        & Who was born first, Aaron Gillespie or Nathan Leone? & \textcolor{red}{Aaron Gillespie was born before Nathan Leone}.   \\
    \midrule
      \multirow{6.5}{*}{\begin{tabular}[x]{@{}l@{}} \textbf{Incompleteness}\\\textbf{Hallucination} \end{tabular}} & List ten book titles on social cognitive theory & \textcolor{red}{1.}``Social Cognitive Theory: An Agentic Perspective'' by Albert Bandura. ... \textcolor{red}{8.}``Cognitive Foundations of Soclal Behavior'' by John L, Hayes and Paul J. Gendolla \\ 
    \cmidrule{2-3}
      & What is the components of an ecosystem? & \textcolor{red}{An ecosystem consists of non-living matter and energy, producers and consumers}, where the producers are mainly plants and the consumers are mainly people, animals, etc. \\
      \midrule
      \multirow{5}{*}{\begin{tabular}[x]{@{}l@{}} \textbf{Outdatedness}\\\textbf{Hallucination} \end{tabular}} & Who is the present president of the United States? & The present president of the United States is \textcolor{red}{Donald Trump}. \\ 
    \cmidrule{2-3}
      & Which team wins the 2022 FIFA World Cup? & \textcolor{red}{France} won the match 4–2, claiming their second World Cup and becoming the fourth consecutive title won by a European team, after Italy in 2006, Spain in 2010, and Germany in 2014. \\
      \midrule
      \multirow{5.5}{*}{\begin{tabular}[x]{@{}l@{}} \textbf{Overclaim}\\\textbf{Hallucination} \end{tabular}} & What is the validity of the assertion that ``TMEM27 is a marker for beta cells.''? & A study suggested that TMEM27 is expressed in purified human beta cells but \textcolor{red}{not in other non-beta pancreatic cells}. \\ 
    \cmidrule{2-3}
      & How to lose weight? & \textcolor{red}{The only way to lose weight is to exercise}, so you need to schedule many fat loss exercises to achieve your weight loss goals. \\
      \midrule
      \multirow{6.5}{*}{\begin{tabular}[x]{@{}l@{}} \textbf{Unverifiability}\\\textbf{Hallucination} \end{tabular}} & Do you agree that ten percent of kidney cancers in the UK are attributable to excess weight? & \textcolor{red}{The World Health Organization has stated that being overweight or obese increases the risk of developing kidney cancer.} \\ 
    \cmidrule{2-3}
      & List ten book titles on social cognitive theory & 1.``Social Cognitive Theory: An Agentic Perspective'' by Albert Bandura. ... 8.``\textcolor{red}{Cognitive Foundations of Soclal Behavior}'' by John L, Hayes and Paul J. Gendolla. \\
    \bottomrule
    \end{tabular}
    \caption{Illustrative examples for six types of LLM hallucinations that we define. We use \textcolor{red}{red} color to denote the incorrect content that contradicts with the world knowledge and unverifiable content that cannot be proven to be true or false by existing available sources.}
    \label{tab:hallu_example}
\end{table*}

%% file: sec/Background.tex
\section{Hallucination in LLMs}
\label{sec-background}
In the field of NLP, \emph{hallucination}  typically refers that the model output contains undesired content that is nonsensical or deviates from the source material~\citep{ji2023survey}, in loose analogy with the phenomenon of hallucination in human psychology~\citep{macpherson2013hallucination}. 


Recently, hallucination has gained prominence alongside the emergence of LLMs  such as ChatGPT.
One prominent feature of LLMs is that they possess rich world knowledge, and can utilize such knowledge to solve various downstream tasks. However, it has been shown that LLMs tend to generate the hallucinatory content, especially under an open-domain setting~\citep{Zhang-CoRR-2023-Siren,Ye-CoRR-2023-Cognitive,Huang-arxiv-2023-A}.   
Despite that hallucination can be defined in different ways, we mainly focus on  \emph{factuality hallucination}, since it has become one of the primary sources of erroneous responses by LLMs. 
In light of this research, we propose a  fine-grained categorization of factuality hallucination in LLMs as follows:

\textbullet~\textbf{Entity-error Hallucination.} This type of hallucination refers to the situations where the generated text of LLMs contains erroneous entities, such as person, date, location, and object, that contradict with the world knowledge. As shown in Table~\ref{tab:hallu_example}, when inquired about ``\textit{the announcement date of Nokia 3510 phone}'', the model generates an erroneous date ``\textit{October 2002}'', in conflict with the real fact ``\textit{March, 2002}''. 

\textbullet~\textbf{Relation-error Hallucination.} This type of hallucination refers to instances where the generated text of LLMs contains wrong relations between entities such as quantitative and chronological relation. As shown in Table~\ref{tab:hallu_example}, the model generates ``\textit{Among the chemical elements that make up the human body, calcium is more than oxygen}'' (wrong quantitative relation) and ``\textit{Aaron Gillespie was born before Nathan Leone}'' (wrong chronological relation).

\textbullet~\textbf{Incompleteness Hallucination.} LLMs might exhibit incomplete output when generating lengthy or listed responses. This hallucination arises when LLMs are asked about aggregated facts and they fail to reserve the factual completeness. For example, as presented in Table~\ref{tab:hallu_example}, when inquired ``\textit{list ten book titles on social cognitive theory}'', the model only generates eight book titles; and the statement ``\textit{in an ecosystem organisms include consumers and producers}'' is factually incomplete.

\textbullet~\textbf{Outdatedness Hallucination.} This type of hallucination refers to situations where the generated content of LLMs is outdated for the present moment, but was correct at some point in the past. This phenomenon arises primarily due to the fact that most LLMs were trained on time-limited  corpora. For example, when asked about ``\textit{the present president of the United States}'', the model trained on corpora before 2021 will generate ``\textit{Donald Trump}'' instead of the latest fact ``\textit{Joe Biden}''. 

\textbullet~\textbf{Overclaim Hallucination.} This type of hallucination refers to cases where the statement expressed in the generated text of LLMs is beyond the scale of factual knowledge~\citep{SchlichtkrullO023}. For example, as shown in Table~\ref{tab:hallu_example}, the model generates overclaimed statements ``\textit{the only way to lose weight is to exercise}'' and ``\textit{you can grow taller just by drinking milk}''. 

\textbullet~\textbf{Unverifiability Hallucination.} In some cases, the information generated by LLMs cannot be verified by available information sources. 
For example, LLMs might  generate plausible but non-existent
academic reading lists. As demonstrated in Table~\ref{tab:hallu_example}, it cannot find any information about the book ``\textit{Cognitive Foundations of Social Behavior}'' from existing sources. 

Note that it is impossible to encompass all kinds of hallucination, and our taxonomy aims to showcase the most frequently occurring types of hallucination.
To make a clear illustration of our categorization of LLM hallucinations, we present more examples for each type in Table~\ref{tab:hallu_example}.

%% file: sec/settings.tex
\section{Experimental Setup}

In this section, we first introduce the construction of our benchmark and then describe a set of comparison models for evaluation.

\subsection{Benchmark Construction}
\label{sec-construction}

To comprehensively evaluate the tendency of LLMs to generate hallucinations across various domains, we extend the previous study of HaluEval~\citep{Li-arxiv-2023-HaluEval} and meticulously construct an upgraded hallucination evaluation benchmark \textbf{HaluEval 2.0}, which contains large-scale questions from five domains, \ie biomedicine, finance, science, education, and open domain. 

Specially, we firstly extract fact-intensive questions from six widely-used domain datasets, including BioASQ~\citep{BioASQ}, NFCorpus~\citep{nfcorpus}, FiQA-2018~\citep{maia201818}, SciFact~\citep{scifact}, LearningQ~\citep{learningq}, and HotpotQA~\citep{hotpotqa}. 
To attain a sufficient number of high-quality questions, we adopt the training set of BioASQ and FiQA-2018, the whole set of NFCorpus and SciFact, and the test set of LearningQ (TED-Ed) and HotpotQA. Then, we only keep input questions ended with the question mark, and for SciFact we transform the input statements into questions using manually designed templates (\eg \textit{[statement] Can you provide some explanations to this statement?}). Finally, following HaluEval~\citep{Li-arxiv-2023-HaluEval} to select questions that LLMs are most likely to hallucinate, we employ ChatGPT to generate three responses for each question and compute their average semantic similarity via BERTScore~\citep{ZhangKWWA20}. 
We only retain those questions where the similarity score is lower than a threshold.
In our benchmark, the questions in biomedicine and science are expert-written, specialized, and require at least high-school knowledge to answer, while the questions in other domains are open-ended, mainly collected from daily conversation and open websites. 
In open domain, the questions from HotpotQA are based on the factual knowledge from Wikipedia.

In total, our HaluEval 2.0 benchmark consists of 8,770 questions from five domains, including 1535, 1125, 1409, 1701, and 3000 questions for biomedicine, finance, science, education, and open domain, respectively.

\subsection{Evaluation Models}

We conduct the experiments with a number of representative open-source and closed-source LLMs based on HaluEval 2.0. 

$\bullet$~\textit{Open-source models.} We mainly focus on instruction-tuned models, which are fine-tuned using instructions (\eg daily chat, synthetic instructions). In our experiments, we select six representative instruction-tuned models including Alpaca (7B)~\citep{alpaca}, Vicuna (7B and 13B)~\citep{vicuna2023}, YuLan-Chat (13B)~\citep{GSAI-github-2023-YuLan}, and Llama 2-Chat (7B and 13B)~\citep{Touvron-2023-llama2-arxiv}. 

$\bullet$~\textit{Closed-source models.} Compared to open-source models, closed-source models can be only accessed via APIs, which usually present exceptional performance. Here, we select five
representative closed-source models including text-davinci-002/003~\citep{Ouyang-arxiv-2022-Training}, ChatGPT, Claude, and Claude 2, where the first three and the latter two models are developed by OpenAI and Anthropic, respectively.

%% file: sec/detection.tex
\section{Hallucination Detection}

To analyze and mitigate hallucinations, the first and fundamental step is to detect hallucinations in model output. In this section, we design an automatic hallucination detection approach and validate its reliability by comparing with human labeling.

\subsection{Detection Approach}

We propose a simple yet effective framework for detecting factual errors in model responses. Following previous work~\citep{Chern-2023-CoRR-FacTool,Dhuliawala-2023-CoRR-Chain}, our approach decomposes the challenging hallucination detection task into two simpler subtasks: 1) extract multiple factual statements from a lengthy response; and 2) determine whether each statement contains hallucinations. Next we will detail the two steps and examine the reliability of our approach.

\paratitle{Fact Extraction.} The responses of LLMs usually contain functional but factually irrelevant content, which hinders the detection of hallucination over the entire complicated responses. Therefore, the first step of our approach is to extract independent factual statements involved in the model response. Instead of training a specific model for fact extraction~\citep{thorne-etal-2018-fact,stanovsky-etal-2018-supervised}, we leverage the strong instruction-following capabilities of LLMs, and instruct GPT-4 to extract factual statements that could be proven to be true or false according to the world knowledge. 
This approach can significantly reduce the costs of data annotation and model training. 
If a response does not contain any factual statements (\eg ``\textit{I don't have enough knowledge to answer your question.}''), the LLM-based fact extractor is allowed to respond ``\textit{NO FACTS}''. We provide several extraction demonstrations to LLMs for in-context learning.

\paratitle{Fact Judgement.} Based on extracted statements, the next step is to determine their trustfulness with respect to the world knowledge. 
Since the LLM has possessed rich world knowledge, we use the LLM (\ie GPT-4) itself to judge the statements.
Previous work checks each involved fact independently using separate prompts based on verification questions or external tools~\citep{Chern-2023-CoRR-FacTool,Dhuliawala-2023-CoRR-Chain}. However, we observe that these statements are often interrelated, with certain statements providing the background or serving as conditions for others. Hence, independently assessing each statement may lead to misjudgment of hallucination. In this work, GPT-4 is instructed with \textit{all} factual statements from the first step to predict their hallucination judgements (\ie \textit{True}, \textit{False}, or \textit{Unknown}). 
To give a confident hallucination judgement, we only consider the \emph{false} statement as hallucination in our following experiments. 


\paratitle{Test of Reliability.} To examine the reliability of our proposed approach in hallucination detection, we invite human labelers to annotate the factuality of {a subset from HaluEval 2.0} and compare the judgement of LLM and humans.
Specially, based on the semantic similarity score (Section~\ref{sec-construction}), we select 1,000 questions from HaluEval 2.0 with an average of 200 samples for each domain. For each question, we use ChatGPT to generate a response and conduct response- and fact-level annotations, where human labelers first annotate whether the response is relevant to the question (\textit{response level}) and then annotate the hallucination type of each factual statement (\textit{fact level}). The hallucination types have been introduced in Section~\ref{sec-background} (Table~\ref{tab:hallu_example}). Note that the fact-level annotation is only conducted when the response is labeled as relevant to the question. To ensure the correctness of our annotation, each sample is labeled by two labelers and examined by one checker. Finally, the matching rate between the judgement of our proposed method and human annotation is 92.6\%, 94.7\%, 92.7\%, 91.5\%, and 93.9\% for biomedicine, finance, science, education, and open domain, respectively. The high consistency demonstrates that our proposed method has a high level of reliability in detecting the hallucinations from LLMs. 

\paratitle{Evaluation Metrics.} Based on our detection approach, we can evaluate a wide range of LLMs on HaluEval 2.0. We design two metrics from different levels to measure the degree of LLMs generating hallucinations in their responses. The \textit{micro hallucination rate (MiHR)} measures the proportion of hallucinatory statements within each response, which is calculated as:
\begin{equation}
   \text{MiHR}  = \frac{1}{n} \sum_{i=1}^{n} \frac{\text{Count}(hallucinatory~facts)}{\text{Count}(all~facts~in~r_i)} , \nonumber
\end{equation}
where $n$ is the total number of samples in every domain and $r_i$ is the $i$-th response. Besides, the \textit{macro hallucination rate (MaHR)} calculates the proportion of responses containing hallucinatory statements, which is computed as: 
\begin{equation}
    \text{MaHR} = \frac{\text{Count}(hallucinatory~responses)}{n} .  \nonumber
\end{equation}
\textbf{For both metrics, smaller value indicates better performance.}  

\begin{table*}[t]
	\small
	\centering
	\begin{tabular}{l c c c c c c c c c c}
		\toprule
		\multicolumn{1}{l}{\multirow{2.5}{*}{\textbf{Models}}} & \multicolumn{2}{c}{\textbf{Biomedicine}} & \multicolumn{2}{c}{\textbf{Finance}} & \multicolumn{2}{c}{\textbf{Science}} & \multicolumn{2}{c}{\textbf{Education}} & \multicolumn{2}{c}{\textbf{Open Domain}} \\ 
		\cmidrule(r){2-3}\cmidrule(r){4-5}\cmidrule(r){6-7}\cmidrule(r){8-9}\cmidrule(r){10-11}
		& MaHR  & MiHR & MaHR  & MiHR & MaHR  & MiHR & MaHR  & MiHR & MaHR  & MiHR \\ 
            \midrule[0.5pt]
            \textbf{ChatGPT} & 14.66 & 3.64 & 25.34 & 6.28 & 18.27 & 4.19 & 33.13 & 8.37 & 47.19 & 13.21 \\
            \textbf{Claude 2} & 28.76 & 7.23 & 35.91 & 9.25 & 15.21 & 3.36 & 36.84 & 10.13 & 39.18* & 12.62*\\
            \textbf{Claude} & 31.44 & 8.25 & 39.11 & 10.56 & 21.31 & 4.78 & 41.26 & 11.53 & 55.39 & 19.50 \\
            \textbf{Text-Davinci-002} & 34.88 & 15.07 & 41.51 & 18.24 & 29.99 & 9.19 & 37.82 & 17.80 & 44.51 & 25.93 \\
            \textbf{Text-Davinci-003} & 46.38 & 14.27 & 56.01 & 16.65 & 43.11 & 12.11 & 58.86 & 19.54 & 70.53 & 25.25\\
            \midrule[0.5pt]
            \textbf{Vicuna 13B} & 50.59 & 17.55 & 46.19 & 13.15 & 34.44 & 8.75 & 55.81 & 17.88 & 65.43 & 29.15 \\
            \textbf{Vicuna 7B} & 52.51 & 18.79 & 50.77 & 14.67 & 40.14 & 10.42 & 58.44 & 19.12 & 66.77 & 29.18 \\
            \textbf{YuLan-Chat 13B} & 60.91 & 22.00 & 46.19 & 14.03 & 41.19 & 10.93 & 52.91 & 17.29 & 68.42 & 30.76 \\
            \textbf{Llama 2-Chat 13B} & 52.61 & 17.90 & 53.48 & 14.53 & 39.11 & 10.37 & 62.12 & 19.30 & 79.19 & 30.44  \\
            \textbf{Llama 2-Chat 7B} & 58.71 & 20.38 & 56.09 & 15.98 & 43.58 & 11.07 & 66.04 & 21.64 & 80.99 & 32.64 \\
            \textbf{Alpaca 7B} & 53.52 & 24.42 & 53.47 & 24.46 & 40.74 & 12.74 & 68.95 & 22.38 & 65.65 & 29.57 \\
            \bottomrule
	\end{tabular}
	\caption{Evaluation results on the tendency of LLMs to generate hallucinations. The lower the hallucination rate, the better the LLM performs. ``*'' represents that Claude 2 always refuses to answer questions in open domain, resulting in a very limited number of valid responses and low hallucination rate.}
	\label{tab:hallucination-rate}
\end{table*}

\begin{table}[t]
	\small
	\centering
	\setlength\tabcolsep{3.3pt}
	\begin{tabular}{l c c c c}
		\toprule
		 \textbf{Models} & \textbf{QA} & \textbf{Dialog} & \textbf{Summari.} & \textbf{Daily Chat} \\
		\midrule
            \textbf{ChatGPT}  & 62.59   & 72.40   & 58.53 & 79.44 \\
            \textbf{Claude 2}  & 69.78   & 64.73   & 57.75 & 75.00 \\
            \textbf{Claude} & 67.60 & 64.83 & 53.76 & 73.88\\
            \textbf{Davinci-003} & 49.65 & 68.37 & 48.07 & 80.40 \\
            \textbf{Davinci-002} & 60.05  & 60.81 & 47.77 & 80.42 \\
		\textbf{GPT-3}    & 49.21 &  50.02  &  51.23 & 72.72 \\
		\midrule
            \textbf{Llama 2-Ch 7B} & 49.60 & 43.99 & 49.55 & 20.46 \\
            \textbf{ChatGLM 6B} & 47.93 & 44.41 & 48.57 & 30.92 \\
            \textbf{Falcon 7B} & 39.66 & 29.08 & 42.71 & 18.98 \\
            \textbf{Vicuna 7B} & 60.34 & 46.35 & 45.62 & 19.48 \\
            \textbf{Alpaca 7B} & 6.68 & 17.55 & 20.63 & 9.54 \\
		\bottomrule
	\end{tabular}
	\caption{Accuracy (\%) on the ability of LLMs to recognize hallucinations for four tasks. The results are copied from HaluEval~\citep{Li-arxiv-2023-HaluEval}.}
	\label{tab:accuracy}
\end{table}

\subsection{Results and Analysis}

We evaluate LLMs on HaluEval 2.0 and apply our detection approach 
to measure their tendency to produce hallucinations in Table~\ref{tab:hallucination-rate}. In addition, we include the results from HaluEval~\citep{Li-arxiv-2023-HaluEval} to showcase the ability of LLMs to recognize hallucinations in Table~\ref{tab:accuracy}.

\paratitle{Tendency to Generate Hallucinations.} From the results in Table~\ref{tab:hallucination-rate}, we can clearly observe that there exists a significant performance gap between open-source and closed-source models. Among open-source models, 
we can discover that scaling the model size can effectively decrease the tendency to generate hallucinations. For example, Vicuna 13B and Llama 2-Chat 13B exhibit much lower hallucination rates than their 7B counterparts. 
Besides, we see that MaHR and MiHR are not strongly positively correlated (\eg Alpaca 7B vs Llama 2-Chat 7B). This is because some models tend to generate shorter responses with fewer facts, which reduce the occurrence of hallucinations but also decrease the richness of information in the replies.
Among closed-source models, after aligning with humans, ChatGPT and Claude 2 show extremely low hallucination rates across five domains. However, we observe that in open domain Claude 2 becomes overly cautious and excessively hedge or ``overrefuse'' innocuous requests by responding ``\textit{Sorry, I don't have enough knowledge to answer this question}'', resulting in a limited number of valid responses and low hallucination rate. 
Furthermore, comparing the results across five domains, it shows that the tendency of LLMs to generate hallucinations is related to specific domains, \ie higher results in education and open domain. Especially in open domain, the hallucination rate of Llama 2-Chat has even reached around 80\%, and ChatGPT and Claude also achieve hallucination rates from 40\% to 50\%.
For open domain, we select the most difficult questions from HotpotQA where ChatGPT is likely to hallucinate. 
These findings suggest that the training methods of current LLMs in open domain are insufficient in preventing from generating hallucinations, while also highlighting the necessity of incorporating domain-specific knowledge into these models. {Note that the percentage results of generating hallucinations in Table~\ref{tab:hallucination-rate} might significantly exceed the actual rate in overall use, because our dataset is specially curated for hallucination evaluation.}

\paratitle{Ability to Recognize Hallucinations.} 
We copy the hallucination detection results in Table~\ref{tab:accuracy} from HaluEval~\citep{Li-arxiv-2023-HaluEval}, which also shows a prominent accuracy gap between open-source and closed-source models. We can see that LLMs are poor at recognizing hallucinations in text. For example, ChatGPT cannot distinguish between factual and hallucinatory summary and only achieves 58.53\% accuracy in summarization, which is barely above chance. Although those open-source models such as Llama 2-Chat and ChatGLM are specially fine-tuned using daily-chat instructions, they have relatively weak abilities in recognizing hallucinations from daily chats. The \emph{open domain} category in HaluEval 2.0 and the \emph{QA task} in HaluEval are both collected from HotpotQA focused on open-domain facts from Wikipedia. We can observe that LLMs, showing higher tendencies to generate hallucinations such as Alpaca, also present limited capabilities to recognize hallucinations. {These results in Table~\ref{tab:hallucination-rate} and Table~\ref{tab:accuracy} 
indicate an implicit correlation between hallucination recognition and generation, but averting from generating hallucinations is more challenging for these models.
}

~

\noindent \emph{\textbf{Note:} The following experiments are conducted on the subset of 1,000 samples from HaluEval 2.0 selected in human annotation.}

%% file: sec/source.tex
\section{Hallucination Source}

\begin{figure*}[t]
	\centering
	\subfigure[Accuracy \textit{w.r.t.} Billions of pretraining tokens]{
		\centering
		\includegraphics[width=0.47\textwidth]{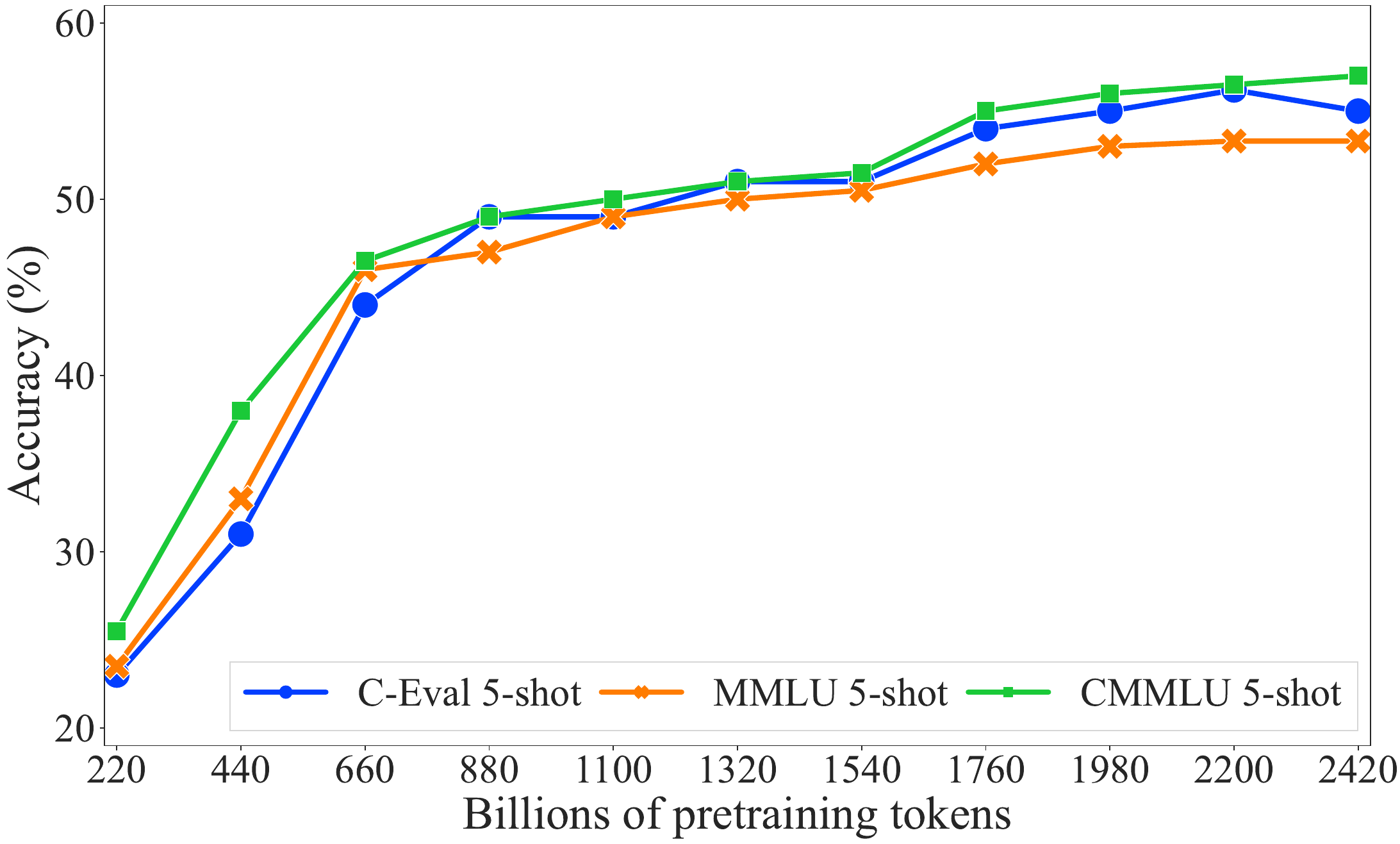}
	}
	\subfigure[Average HR \textit{w.r.t.} Billions of pretraining tokens]{
		\centering
		\includegraphics[width=0.47\textwidth]{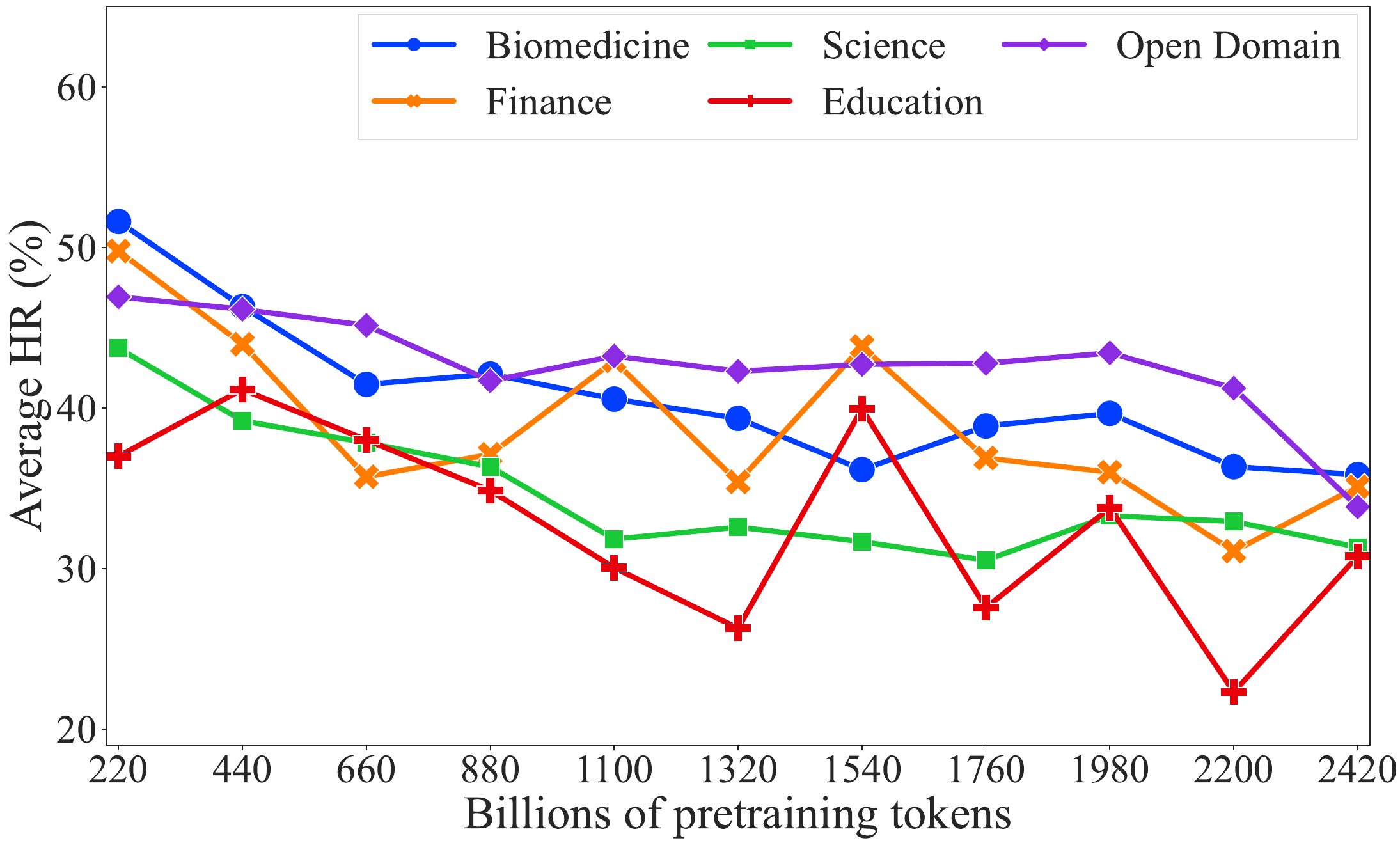}
	}
	\centering
	\caption{Baichuan 2 (7B) task accuracy (\%) in three benchmarks and average hallucination rate (\%) in five domains with respect to billions of pretraining tokens. We average the macro and micro hallucination rates. The  accuracy results in (a) are copied from Baichuan 2~\citep{baichuan2}.}
	\label{fig:pre-training-tokens}
	\vspace{-0.2cm}
\end{figure*}

\begin{table*}[t]
	\small
	\centering
	\begin{tabular}{l c c c c c c c c c c}
		\toprule
		\multicolumn{1}{l}{\multirow{2.5}{*}{\textbf{Models}}} & \multicolumn{2}{c}{\textbf{Biomedicine}} & \multicolumn{2}{c}{\textbf{Finance}} & \multicolumn{2}{c}{\textbf{Science}} & \multicolumn{2}{c}{\textbf{Education}} & \multicolumn{2}{c}{\textbf{Open Domain}} \\ 
		\cmidrule(r){2-3}\cmidrule(r){4-5}\cmidrule(r){6-7}\cmidrule(r){8-9}\cmidrule(r){10-11}
		& MaHR  & MiHR & MaHR  & MiHR & MaHR  & MiHR & MaHR  & MiHR & MaHR  & MiHR \\ 
            \midrule[0.5pt]
            \textbf{Falcon 40B} & 53.25 & 20.25 & 39.61 & 15.35 & 36.75 & 10.89 & 47.83 & 22.82 & 61.96 & 28.94 \\
            \textbf{Galactica 30B} & 51.70 & 16.51 & 58.93 & 23.18 & 41.72 & 11.24 & 51.76 & 18.09 & 52.17 & 25.25 \\
            \textbf{GPT-NeoX 20B} & 63.27 & 22.29 & 60.58 & 23.38 & 46.95 & 14.28 & 61.17 & 23.77 & 61.62 & 29.26 \\
            \bottomrule
	\end{tabular}
	\caption{Evaluation results in five domains for three pre-trained models with diverse pre-training corpus.}
        \vspace{-0.1cm}
	\label{tab:mixture-results}
\end{table*}

In this section, we perform a series of experiments to explore four factors that may induce LLM hallucinations, including pre-training, supervised fine-tuning, prompt design, and inference methods. 

\subsection{Pre-training}

Pre-training serves as the fundamental  stage to establish the abilities of LLMs, enabling them to gain general capabilities and rich world knowledge. However, carrying out a detailed investigation of all possible hallucination factors in pre-training is infeasible in practice, due to the high training cost from scratch and the undisclosed technical details in existing LLMs.   
In this part, we conduct the study by considering two key factors, namely \emph{scale} and \emph{source} of pre-training data. We select these two factors because they are relatively easy to be examined according to public disclosure of LLMs and often have a large impact on model performance. 
Specially, we explore the effects of pre-training on LLM hallucinations in three aspects related to the scale and source of pre-training data.   


\begin{figure}[t]
    \centering
    \includegraphics[width=0.46\textwidth]{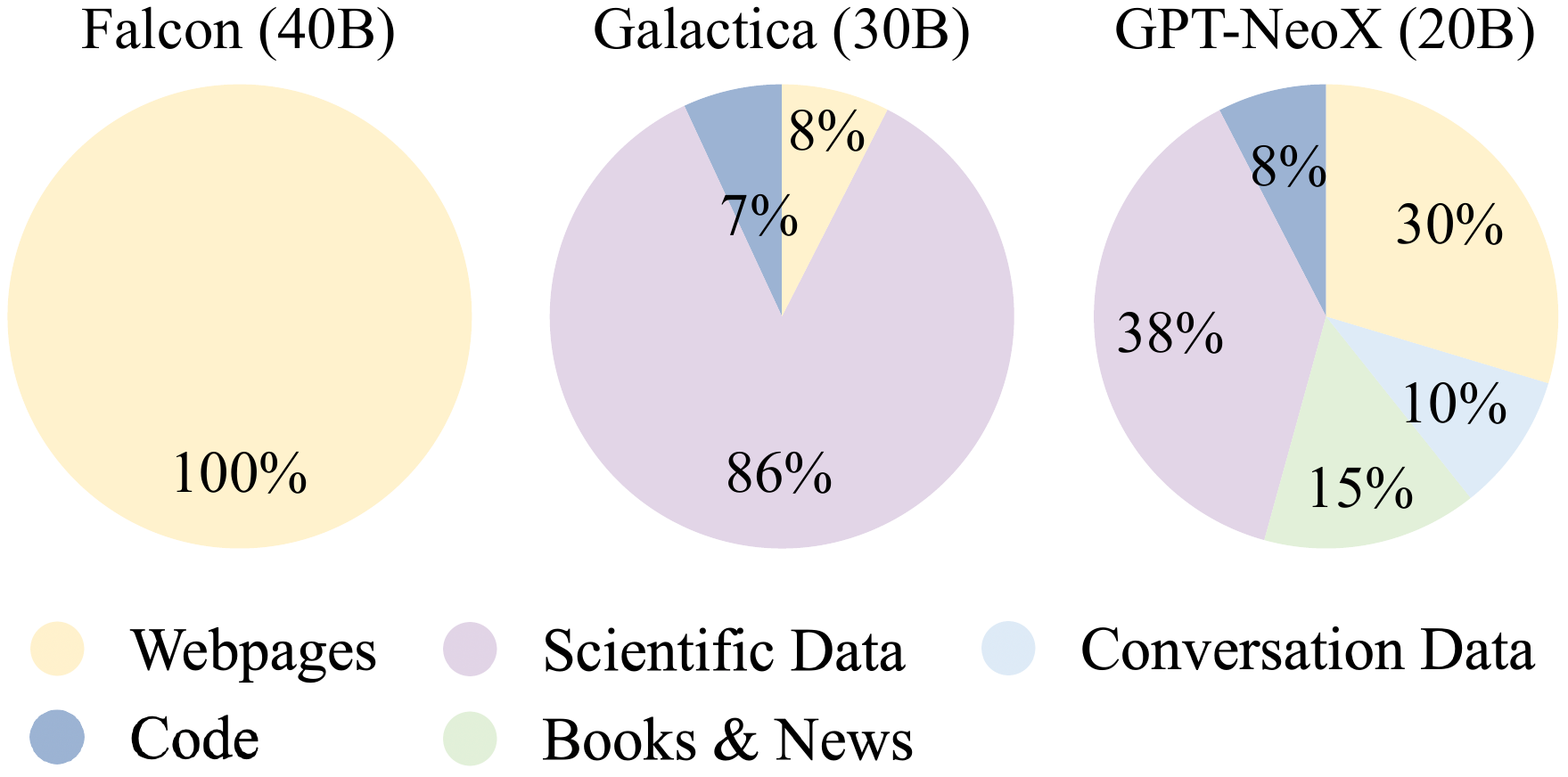}
    \caption{Ratios of pre-training data sources (figure copied from the LLM survey article~\citep{zhao2023survey}).}
    \vspace{-0.2cm}
    \label{fig:corpus_mixture}
\end{figure}

\paratitle{Amount of Pre-training Tokens.} 
We select 11 intermediate pre-training checkpoints of Baichuan 2 (7B)~\citep{baichuan2}, corresponding to training on approximately 0.2 to 2.4 trillion tokens, and evaluate them on three commonly-used benchmarks and HaluEval 2.0. As shown in Figure~\ref{fig:pre-training-tokens}, with training on an increasing amount of tokens, Baichuan 2 demonstrates  improved general task performance in benchmarks. However, the hallucination rate oscillates across these model checkpoints, suffering more in the domains of finance and education while overall decreasing in the domains of biomedicine, science and open domain.
This finding indicates that simply increasing pre-training tokens may not be that effective in hallucination reduction, which may require specific data strategies to alleviate the hallucinations in some specific domains. 

\paratitle{Mixture of Pre-training Corpus.} Existing LLMs typically employ a mixture of diverse public textual datasets as the pre-training corpus. 
The distribution of data source will affect the acquisition of general and domain-specific knowledge. To examine the effect of pre-training data mixture on hallucinations, we evaluate three LLMs with similar model sizes but using different pre-training corpus of general data (\eg webpages) and specialized data (\eg scientific text): Falcon (40B)~\citep{Ebtesam-arxiv-2023-Falcon}, Galactica (30B)~\citep{Taylor-arxiv-2022-Galactica}, and GPT-NeoX (20B)~\citep{Black-CoRR-2022-GPT}. Their pre-training data distribution is shown in Figure~\ref{fig:corpus_mixture} and our results are shown in Table~\ref{tab:mixture-results}. 
We can see that training on scientific data can significantly prevent the model from generating hallucinations in science and open domain (Galactica v.s. GPT-NeoX). 
General webpages data benefits reducing hallucinations in the domains of finance, science, and education.
On the other hand, training on diverse corpora (\ie GPT-NeoX) instead tends to result in much more hallucinations. Note that a potential factor to affect our results is that the data cleaning procedure for each model, since it would produce pre-training data of different data qualities. We leave this discussion of  data cleaning in future work. 

\begin{figure*}[t]
	\centering
	\subfigure[ChatGPT]{
		\centering
		\includegraphics[width=0.47\textwidth]{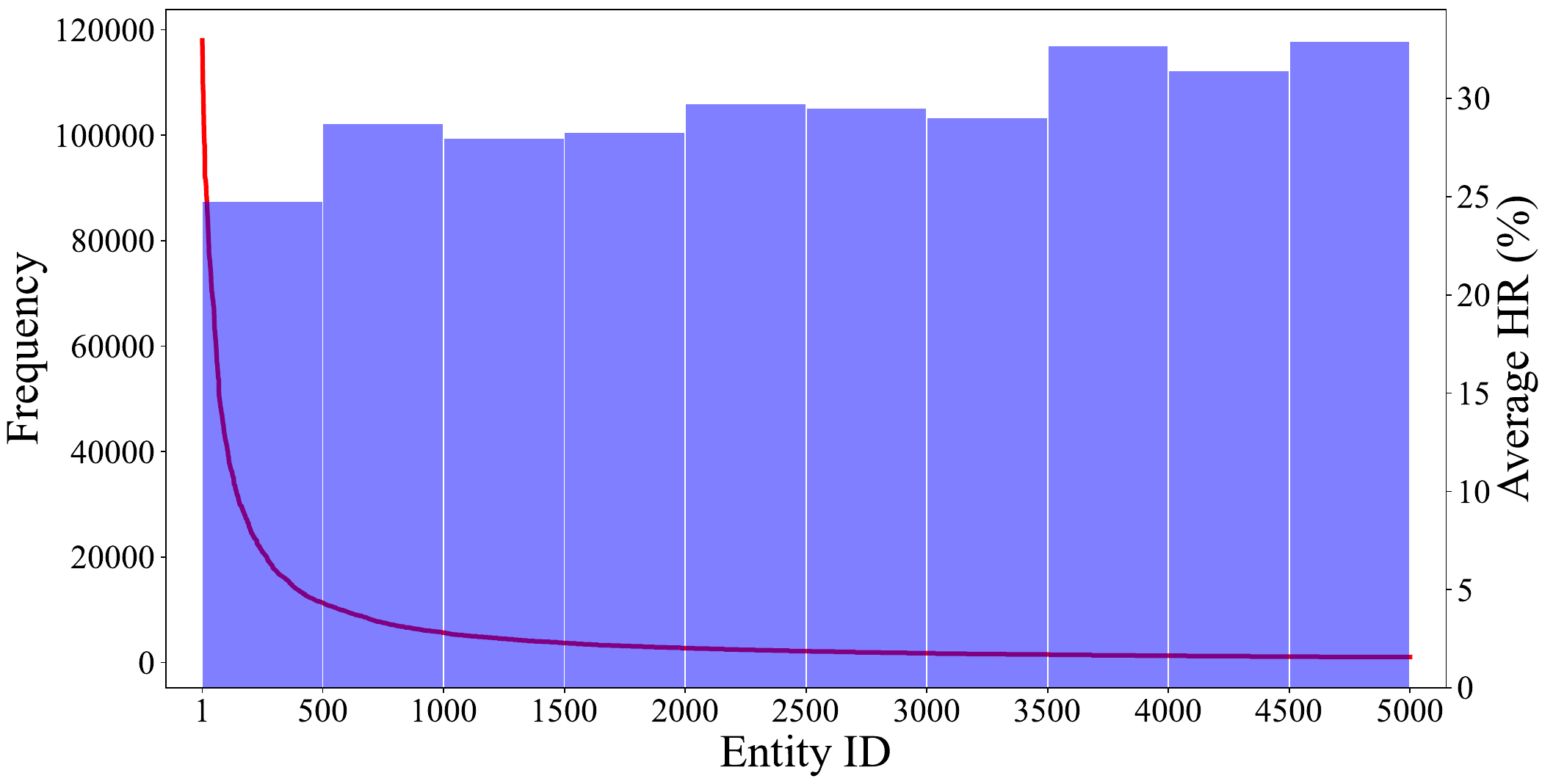}
	}
	\subfigure[Llama 2-Chat (7B)]{
		\centering
		\includegraphics[width=0.47\textwidth]{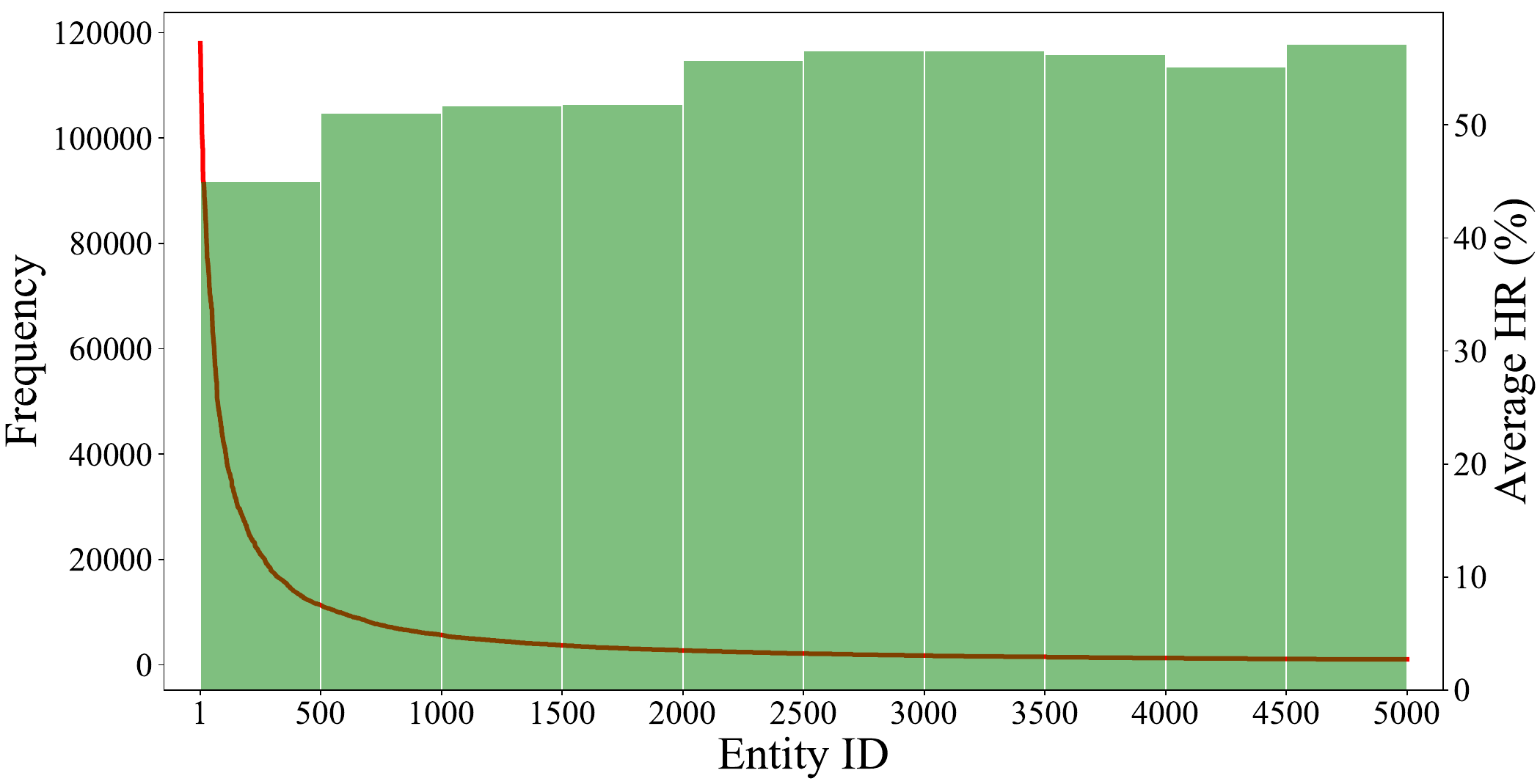}
	}
	\centering
	\caption{Evaluation results of ChatGPT and Llama 2-Chat (7B). The \textcolor{red}{red} line denotes the frequency of entities, and the \textcolor{blue}{blue} and \textcolor{green}{green} bar denotes the average hallucination rate (\%) for each group of entities.}
	\label{fig:entity_frequency}
\end{figure*}

\paratitle{Familiarity of Pre-training Knowledge.} Previous work has reported that LLMs tend to produce hallucinations for those infrequent knowledge facts in pre-training corpus~\citep{LiLSDSLJJL22}. To investigate this, we adopt the  commonly used Wikipedia~\citep{dpr2020} and collect the entity set based on the document titles. Then, we calculate the occurrence frequencies of entities in the whole document corpus and keep top $5,000$ entities with the highest frequencies and categorize these entities into $10$ groups with descending frequencies. 
Finally, we randomly choose $1,000$ entities (with an average of $100$ entities from each group) and construct a set of entity-based questions using manually designed templates (\eg ``\textit{please offer an introduction to [entity]}''). 
We evaluate ChatGPT and Llama 2-Chat (7B) on these questions.
As shown in Figure~\ref{fig:entity_frequency}, we can clearly observe a long-tail distribution on the entity frequency. {Interestingly, the hallucination rates exhibit a clear three-level stair-like pattern on the 10 entity groups. ChatGPT and Llama 2-Chat (7B) show the lowest tendency to generate hallucinations for those entities in the first group, which have over 85\% occurrence frequencies but only occupy 10\% of the entire set of entities. For the vast majority of entities in the long tail of frequency, the model exhibits a relative high hallucination rate.}

\begin{table*}[t]
	\small
	\centering
	\begin{tabular}{l c c c c c c c c c c}
		\toprule
		\multicolumn{1}{l}{\multirow{2.5}{*}{\textbf{Datasets}}} & \multicolumn{2}{c}{\textbf{Biomedicine}} & \multicolumn{2}{c}{\textbf{Finance}} & \multicolumn{2}{c}{\textbf{Science}} & \multicolumn{2}{c}{\textbf{Education}} & \multicolumn{2}{c}{\textbf{Open Domain}} \\ 
		\cmidrule(r){2-3}\cmidrule(r){4-5}\cmidrule(r){6-7}\cmidrule(r){8-9}\cmidrule(r){10-11}
		& MaHR  & MiHR & MaHR  & MiHR & MaHR  & MiHR & MaHR  & MiHR & MaHR  & MiHR \\ 
            \midrule[0.5pt]
            \ding{172}~FLAN-T5 & 73.12 & 32.33 & 70.29 & 28.79 & 45.00 & 18.33 & 67.55 & 32.73 & 64.44 & 31.48 \\
            \ding{173}~ShareGPT & 66.11 & 25.34 & 68.25 & 23.92 & 42.21 & 11.66 & 66.67 & 25.00 & 62.20 & 28.33 \\
            \ding{174}~Self-Instruct-52K & 71.11 & 31.33 & 69.65 & 30.82 & 43.52 & 14.69 & 67.00 & 30.88 & 63.62 & 30.56 \\
            \ding{173} + \ding{174} & 70.17 & 29.60 & 67.22 & 24.02 & 39.69 & 12.52 & 66.94 & 27.76 & 56.56 & 29.43 \\
            \ding{172} + \ding{173} + \ding{174} & 64.52 & 26.80 & 70.00 & 25.29 & 43.81 & 13.56 & 67.14 & 28.63 & 56.67 & 31.56 \\
            \midrule[0.5pt]
            \ding{174}~Self-Instruct-52K & 71.11 & 31.33 & 69.65 & 30.82 & 43.52 & 14.69 & 67.00 & 30.88 & 63.62 & 30.56 \\
            w/ complexity & 69.74 & 31.08 & 62.70 & 21.17 & 42.21 & 13.02 & 69.34 & 30.21 & 55.40 & 29.34 \\
            w/ diversity & 67.71 & 27.45 & 68.45 & 26.22 & 41.21 & 12.36 & 64.00 & 26.31 & 63.10 & 30.40 \\
            w/ scaling & 65.98 & 27.80 & 67.63 & 29.24 & 37.50 & 12.42 & 64.10 & 29.59 & 53.33 & 29.96 \\
            w/ difficulty & 57.38 & 25.78 & 65.87 & 30.86 & 31.47 & 12.20 & 49.62 & 27.89 & 42.96 & 24.06 \\
            \bottomrule
	\end{tabular}
	\caption{Evaluation results of LLaMA (7B) after supervised fine-tuning with different mixture of instruction datasets and types of instruction synthesis.}
        \vspace{-0.1cm}
	\label{tab:sft-results}
\end{table*}

\begin{tcolorbox}
[colback=red!5!white,colframe=red!55!black,title=\textbf{\footnotesize \textsc{\ul{Effects of Pre-training on llm hallucinations}}:}]
\begin{itemize}
[leftmargin=1mm]
\setlength\itemsep{0em}
    \item[\ding{224}] {\footnotesize 
    {\fontfamily{phv}\fontsize{8}{9}\selectfont
   Pre-training on more tokens benefits the general task performance, but has an oscillatory effect on reducing the generation of LLM hallucinations. }} 
    
    \item[\ding{224}] {\footnotesize 
    {\fontfamily{phv}\fontsize{8}{9}\selectfont
   Incorporating specialized data into pre-training corpora can extremely reduce the occurrence of hallucination in specific domains.}}

     \item[\ding{224}] {\footnotesize 
    {\fontfamily{phv}\fontsize{8}{9}\selectfont
     The frequency of pre-training knowledge largely influences the occurrence of LLM hallucinations, \ie lower frequency, more hallucinations.
    }}
\end{itemize}
\end{tcolorbox}

\subsection{Supervised Fine-Tuning}

Supervised fine-tuning involves training language models to follow natural language  instructions provided by the user (\ie prompt). {However, training LLMs on instructions they haven't ever seen the relevant knowledge in pre-training will encourage them to produce hallucinations~\citep{John-youtube-2023-RLHF}.} 
Following the experimental setup in the LLM survey~\citep{zhao2023survey}, we investigate the effect of instructions by mixing different types of instruction datasets and synthesizing new instructions. 

\paratitle{Mixture of Instruction Datasets.} Generally, there are three kinds of instructions, \ie task-specific, daily chat, and synthetic instructions. Hence, we select three representative instruction datasets, including FLAN-T5~\citep{Chung-arxiv-2022-Scaling}, ShareGPT~\citep{ShareGPT}, and Self-Instruct-52K~\citep{Wang-arXiv-2022-Self}, and then fine-tune LLaMA (7B) on each individual instruction set and their mixture to examine combinatorial effects. For a fair comparison we randomly sample $40,000$ instructions for each instruction dataset.
As can be seen from Table~\ref{tab:sft-results}, compared to other instructions, daily chat instructions result in a lower level of hallucinations, while task-specific instructions lead to a higher proportion of hallucinatory responses. The task-specific instructions mainly focus on task format learning and ignore the factual knowledge.
Mixing the daily chat and synthetic instructions can reduce the emergence of LLM hallucinations in some domains such as finance and science. 


\paratitle{Types of Instruction Synthesis.} 
To improve the capacities of LLMs, existing studies typically design a series of strategies to automatically construct large-scale instruction tuning data, \eg data scaling and diversity increasing. Here, we would like to investigate the effect of these instruction synthesis strategies on model hallucination. 
Specially, we consider examining four instruction synthesis strategies based on Self-Instruct-52K: (1) \emph{Enhancing the instruction complexity:} we adopt 40K instructions from WizardLM~\citep{Xu-arxiv-2023-WizardLM} that increases the complexity level by adding constraints, increasing reasoning steps, and complicating the input; (2) \emph{Increasing the topic diversity:} we use ChatGPT to rewrite instructions for adapting to 293 topics and obtain 40K instructions; (3) \emph{Scaling the instruction number:} we sample new instructions from
MOSS~\citep{sun2023moss} and mix them with Self-Instruct-52K to obtain 100K instructions; and (4) \emph{Balancing the instruction difficulty:} we compute the perplexity of instructions by LLaMA (7B) to estimate their difficulty and remove too easy or too hard instructions from the above 100K instructions and only keep 40K instructions. The results are shown in Table~\ref{tab:sft-results}. We can clearly observe that fine-tuning LLMs with improved instructions can be useful to alleviate the phenomenon of  hallucinations. Balancing the complexity of instructions can significantly reduce the generation of hallucinations, while overly complex instructions eventually result in a higher level of hallucinations. 

\begin{table*}[t]
	\small
	\centering
	\begin{tabular}{l c c c c c c c c c c}
		\toprule
		\multicolumn{1}{l}{\multirow{2.5}{*}{\textbf{Models}}} & \multicolumn{2}{c}{\textbf{Biomedicine}} & \multicolumn{2}{c}{\textbf{Finance}} & \multicolumn{2}{c}{\textbf{Science}} & \multicolumn{2}{c}{\textbf{Education}} & \multicolumn{2}{c}{\textbf{Open Domain}} \\ 
		\cmidrule(r){2-3}\cmidrule(r){4-5}\cmidrule(r){6-7}\cmidrule(r){8-9}\cmidrule(r){10-11}
		& MaHR  & MiHR & MaHR  & MiHR & MaHR  & MiHR & MaHR  & MiHR & MaHR  & MiHR \\ 
            \midrule[0.5pt]
            \textbf{ChatGPT} & \\
            w/ base prompt & 48.75 & 14.03 & 46.84 & 13.55 & 24.14 & 6.39 & 53.44 & 17.19 & 59.77 & 17.93 \\
            + {manual desc} & 45.64 & 13.91 & 39.20 & 11.18 & 22.34 & 5.28 & 55.68 & 17.73 & 64.52 & 20.31 \\
            + {synthetic desc} & 51.00 & 14.23 & 44.33 & 11.93 & 25.00 & 6.15 & 55.87 & 17.79 & 52.02 & 17.20 \\
            + {refined question} & 50.76 & 14.89 & 44.16 & 12.11 & 25.25 & 6.29 & 53.19 & 16.27 & 64.36 & 20.27 \\
            + {manual demo} & 42.71 & 14.89 & 40.74 & 12.12 & 25.27 & 7.11 & 56.41 & 19.24 & 44.72 & 21.88 \\
            + retrieved demo & 46.52 & 16.02 & 42.78 & 11.62 & 19.59 & 4.87 & 51.25 & 18.86 & 50.34 & 20.66 \\
            + synthetic demo & 38.10 & 18.30 & 36.69 & 15.13 & 27.71 & 8.15 & 43.90 & 23.24 & 29.17 & 18.08 \\
            + {reverse position} & 54.82 & 15.67 & 48.22 & 13.98 & 26.77 & 6.67 & 51.60 & 16.94 & 67.21 & 21.10 \\
            \midrule[0.5pt]
            \textbf{Llama 2-Chat 7B} &  \\
            w/ {base prompt} & 69.12 & 26.69 & 69.41 & 24.59 & 49.25 & 14.05 & 71.52 & 27.74 & 77.35 & 33.15 \\
            + {manual desc} & 68.02 & 26.46 & 74.36 & 25.01 & 42.50 & 12.10 & 76.16 & 30.97 & 79.39 & 33.23 \\
            + {synthetic desc} & 75.25 & 29.56 & 66.33 & 23.27 & 41.00 & 12.02 & 72.16 & 29.31 & 78.45 & 34.50 \\
            + {refined question} & 74.87 & 31.35 & 68.02 & 25.04 & 44.00 & 13.22 & 72.83 & 29.50 & 81.92 & 35.11 \\
            + {manual demo} & 69.70 & 27.90 & 66.33 & 24.61 & 45.00 & 12.27 & 71.01 & 27.02 & 66.88 & 31.84 \\
            + retrieved demo & 66.33 & 26.94 & 72.36 & 26.31 & 42.50 & 13.19 & 70.06 & 29.76 & 67.24 & 35.56 \\
            + synthetic demo & 59.68 & 27.31 & 62.84 & 24.43 & 45.50 & 14.72 & 57.64 & 27.49 & 53.77 & 30.27 \\
            + {reverse position} & 70.92 & 29.52 & 75.39 & 26.35 & 41.00 & 12.22 & 71.51 & 29.40 & 73.89 & 32.29 \\
            \bottomrule
	\end{tabular}
	\caption{Evaluation results of ChatGPT and Llama 2-Chat (7B) using different prompt formats. }
	\label{tab:prompt_format}
\end{table*}

\begin{tcolorbox}
[colback=red!5!white,colframe=red!55!black,title=\textbf{\footnotesize \textsc{\ul{Effects of Supervised Fine-Tuning on llm hallucinations}}:}]
\begin{itemize}
[leftmargin=1mm]
\setlength\itemsep{0em}
    \item[\ding{224}] {\footnotesize 
    {\fontfamily{phv}\fontsize{8}{9}\selectfont
   The daily-chat instructions result in a lower level of hallucinations, while task-specific instructions lead to a higher proportion of hallucinatory responses. Mixing the daily chat and synthetic instructions can reduce the emergence of LLM hallucinations in some domains such as finance and science.}} 
    
    \item[\ding{224}] {\footnotesize 
    {\fontfamily{phv}\fontsize{8}{9}\selectfont
   Fine-tuning LLMs with improved instructions can be useful to alleviate the phenomenon of hallucinations. Balancing the complexity of instructions can significantly reduce the generation of hallucinations, while using overly complex and diverse instructions results in a higher level of hallucinations.}}
\end{itemize}
\end{tcolorbox}

\subsection{Prompt Design}\label{sec:prompt_design}

Prompting has become the major approach to utilizing LLMs. However, inappropriate prompt design would lead to incapable attention of important information in the input~\citep{Liu-2023-arxiv-Lost}. In addition, ambiguous and superficial questions posed by users might steer the model towards generating unrelated, implausible, or bizarre output~\citep{Rawte-2023-arxiv-Exploring}. In this part, we continue to analyze the effect of prompt design on LLM hallucinations. 

\paratitle{Prompt Design.} Generally, a prompt contains task description, input question, and contextual information such as in-context demonstrations~\citep{SantuF23}. Here, we experiment with several prompt designs by varying the three ingredients:

$\bullet$~\emph{Base prompt}: the initial prompt with a simple task description and input question.

$\bullet$~\emph{Manual description prompt}: manually rewriting the task description in base prompt.

$\bullet$~\emph{Synthetic description prompt}: using ChatGPT to synthesize the task description in base prompt.

$\bullet$~\emph{Refined question prompt}: refining the initial question in base prompt by ChatGPT.

$\bullet$~\emph{Manual in-context prompt}: manually selecting five in-context demonstrations for the base prompt.

$\bullet$~\emph{Retrieved in-context prompt}: retrieving demonstrations based on BERT similarity from HaluEval 2.0 (besides the 1,000 test samples). 

$\bullet$~\emph{Synthetic in-context prompt}: using ChatGPT to synthesize the in-context demonstrations.


$\bullet$~\emph{Reverse prompt}: reversing the position of task description and input question in base prompt (\ie place the task description after the input question).

\noindent We feed these prompts into ChatGPT and Llama 2-Chat (7B) and the evaluation results are shown in Table~\ref{tab:prompt_format}. {First, we can observe that rewriting the task description with more details can reduce the hallucinations to some extent, while this effect is varied in domains. For professional domains (\ie biomedicine and science), incorporating more details into the task description can mitigate some hallucinations.
Second, leveraging in-context learning can also help eliminate hallucinations in LLM's responses. These examplars or demonstrations can be manually selected, retrieved from candidate corpus, or automatically generated by LLM itself. It is noting that MaHR and MiHR are not strongly positively correlated, where the two metrics measure the hallucination degree from distinct levels.
Finally, in most cases, rewriting the question or placing the task description at the end of the input question will instead hurt the model performance and induces more hallucinations.}




\begin{figure*}[t]
    \centering
    \includegraphics[width=\textwidth]{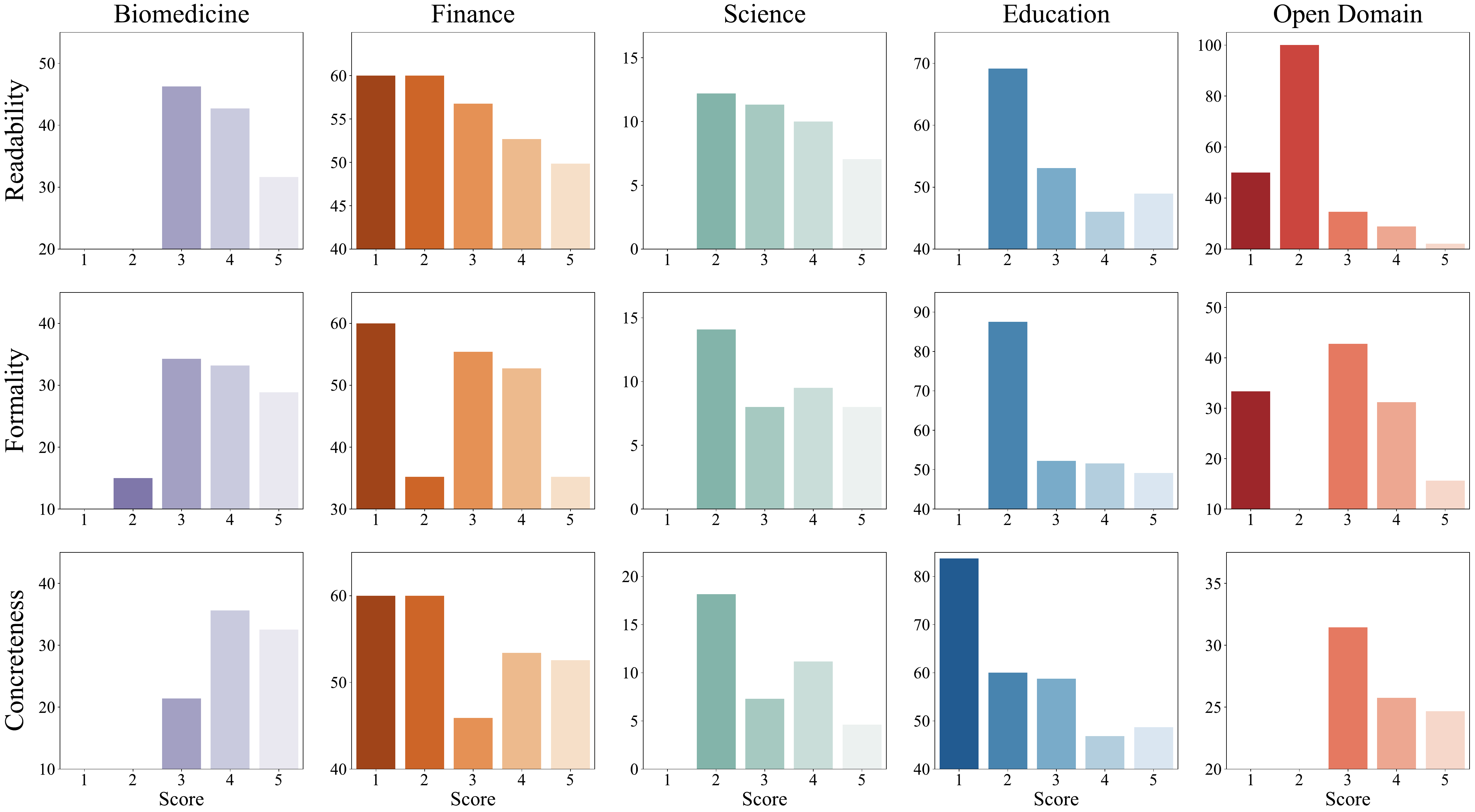}
    \caption{The average hallucination rate (\%) of those responses and questions by ChatGPT for each score of the three properties, \ie readability, formality, and concreteness, in five domains. Some values are zero because there are no scores from humans. The 5-point denotes ``very satisfying'' and the 1-point denotes ``very terrible''.
    }
    \vspace{-0.3cm}
    \label{fig:question_content}
\end{figure*}

\paratitle{Question Content.} Following prior work~\citep{Rawte-2023-arxiv-Exploring}, we further delve into how linguistics of question content, specifically readability, formality, and concreteness, influence the occurrence of LLM hallucinations. \emph{Readability} quantifies the extent to which the question can be understood by humans; \emph{Formality} refers to the degree of appropriate tone and professionalism conveyed by the choice of words, grammatical structure, style, etc.; \emph{Concreteness} indicates whether a word represents a specific and tangible concept. For each question in our dataset, we invite human labelers to score these three properties based on the 5-point Likert scale, ranging from 1-point (``very terrible'') to 5-point (``very satisfying''). For these questions, we instruct ChatGPT to generate responses and compute the average hallucination rate under each score. In Figure~\ref{fig:question_content}, we can observe that ChatGPT exhibits a lower propensity to generate hallucinations on those questions that are easier to read and use more formal and specific language. Note that constrained by the scale of the annotation dataset, there were fewer/no scores from humans for certain linguistics (\eg ``2-point'' formality in biomedicine), resulting in a relatively low hallucination rate.

\begin{tcolorbox}
[colback=red!5!white,colframe=red!55!black,title=\textbf{\footnotesize \textsc{\ul{Effects of Prompt Design on llm hallucinations}}:}]
\begin{itemize}
[leftmargin=1mm]
\setlength\itemsep{0em}
    
    \item[\ding{224}] {\footnotesize 
    {\fontfamily{phv}\fontsize{8}{9}\selectfont
   Incorporating more details into task description can reduce hallucinations, especially in professional domains. Leveraging in-context learning is also helpful. Rewriting the question or placing task description at the end of the question induces more hallucinations.}}

     \item[\ding{224}] {\footnotesize 
    {\fontfamily{phv}\fontsize{8}{9}\selectfont
     LLMs demonstrate a lower propensity to produce hallucinations on those questions that are easier to read and use more formal and specific language. 
    }}
\end{itemize}
\end{tcolorbox}

\begin{table*}[t]
	\small
	\centering
	\begin{tabular}{l c c c c c c c c c c}
		\toprule
		\multicolumn{1}{l}{\multirow{2.5}{*}{\textbf{Models}}} & \multicolumn{2}{c}{\textbf{Biomedicine}} & \multicolumn{2}{c}{\textbf{Finance}} & \multicolumn{2}{c}{\textbf{Science}} & \multicolumn{2}{c}{\textbf{Education}} & \multicolumn{2}{c}{\textbf{Open Domain}} \\ 
		\cmidrule(r){2-3}\cmidrule(r){4-5}\cmidrule(r){6-7}\cmidrule(r){8-9}\cmidrule(r){10-11}
		& MaHR  & MiHR & MaHR  & MiHR & MaHR  & MiHR & MaHR  & MiHR & MaHR  & MiHR \\ 
            \midrule[0.5pt]
            \textbf{ChatGPT} &  \\
            w/ greedy search & 48.75 & 14.03 & 46.84 & 13.55 & 24.14 & 6.39 & 53.44 & 17.19 & 59.77 & 17.93 \\
            w/ top-$p$ sampling & 49.50 & 14.56 & 44.22 & 13.16 & 24.60 & 6.98 & 47.19 & 15.57 & 59.90 & 19.11 \\
            \midrule[0.5pt]
            \textbf{Llama 2-Chat 7B} &  \\
            w/ greedy search & 69.12 & 26.69 & 69.41 & 24.59 & 49.25 & 14.05 & 71.52 & 27.74 & 77.35 & 33.15 \\
            w/ top-$p$ sampling & 79.70 & 34.48 & 63.36 & 21.25 & 50.20 & 15.13 & 61.45 & 25.47 & 74.86 & 30.70 \\
            w/ top-$k$ sampling & 76.33 & 34.04 & 59.43 & 19.81 & 50.23 & 15.78 & 63.99 & 27.16 & 72.26 & 31.00 \\
            w/ beam search & 73.13 & 32.47 & 59.54 & 21.06 & 47.00 & 14.55 & 63.35 & 25.24 & 73.84 & 29.22 \\
            \bottomrule
	\end{tabular}
	\caption{Evaluation results of different decoding strategies for ChatGPT and Llama 2-Chat (7B). Due to the API constraint of ChatGPT, we only investigate greedy search and top-$p$ sampling.}
	\label{tab:decoding_strategies}
\end{table*}

\begin{table*}[t]
	\small
	\centering
	\begin{tabular}{l c c c c c c c c c c}
		\toprule
		\multicolumn{1}{l}{\multirow{2.5}{*}{\textbf{Models}}} & \multicolumn{2}{c}{\textbf{Biomedicine}} & \multicolumn{2}{c}{\textbf{Finance}} & \multicolumn{2}{c}{\textbf{Science}} & \multicolumn{2}{c}{\textbf{Education}} & \multicolumn{2}{c}{\textbf{Open Domain}} \\ 
		\cmidrule(r){2-3}\cmidrule(r){4-5}\cmidrule(r){6-7}\cmidrule(r){8-9}\cmidrule(r){10-11}
		& MaHR  & MiHR & MaHR  & MiHR & MaHR  & MiHR & MaHR  & MiHR & MaHR  & MiHR \\ 
            \midrule[0.5pt]
            \textbf{Llama 2-Chat 7B} &  \\
            INT 16 (original) & 69.12 & 26.69 & 69.41 & 24.59 & 49.25 & 14.05 & 71.52 & 27.74 & 77.35 & 33.15 \\
            
            INT 8 & 76.84 & 33.96 & 70.21 & 23.13 & 50.50 & 13.80 & 78.53 & 26.77 & 73.73 & 30.36 \\
            
            INT 4 & 76.16 & 31.80 & 67.54 & 23.08 & 48.50 & 14.30 & 79.97 & 30.00 & 79.41 & 35.00 \\
            \midrule[0.5pt]
            \textbf{Llama 2-Chat 13B} &  \\
            INT 16 (original) & 70.56 & 26.63 & 69.95 & 23.85 & 42.21 & 13.02 & 69.14 & 26.71 & 76.34 & 32.48 \\
            
            INT 8 & 75.66 & 32.43 & 69.53 & 23.77 & 45.20 & 14.38 & 69.94 & 26.74 & 78.57 & 33.70 \\
            
            INT 4 & 77.96 & 33.84 & 68.49 & 23.64 & 44.20 & 14.21 & 74.71 & 27.70 & 79.12 & 33.71 \\
            \bottomrule
	\end{tabular}
	\caption{Evaluation results for quantized Llama 2-Chat (7B and 13B). }
        \vspace{-0.1cm}
	\label{tab:quantization}
\end{table*}



\subsection{Inference Methods}\label{sec:inference_method}

In the inference stage, special decoding strategies can be used to enhance the generation diversity, such as top-$p$ sampling, but likely contribute to increased  hallucinations~\citep{DziriMZB21}. Besides, most LLMs generate responses in a sequential token-by-token manner, which makes them over-commit to early generation mistakes~\citep{Azaria-2023-arxiv-The}, known as \emph{hallucination snowballing}~\citep{Zhang-2023-arxiv-How}. Furthermore, to reduce the inference latency, quantization is proposed to compress the large model copy in space~\citep{Gholami-CoRR-2022-A}, which instead would sacrifice the model performance and might lead to the generation of hallucinations.
In this part, we will systematically study the effects of these inference related methods on model hallucination. 

\begin{figure}[t]
	\centering
	\subfigure[ChatGPT]{
		\centering
		\includegraphics[width=0.22\textwidth]{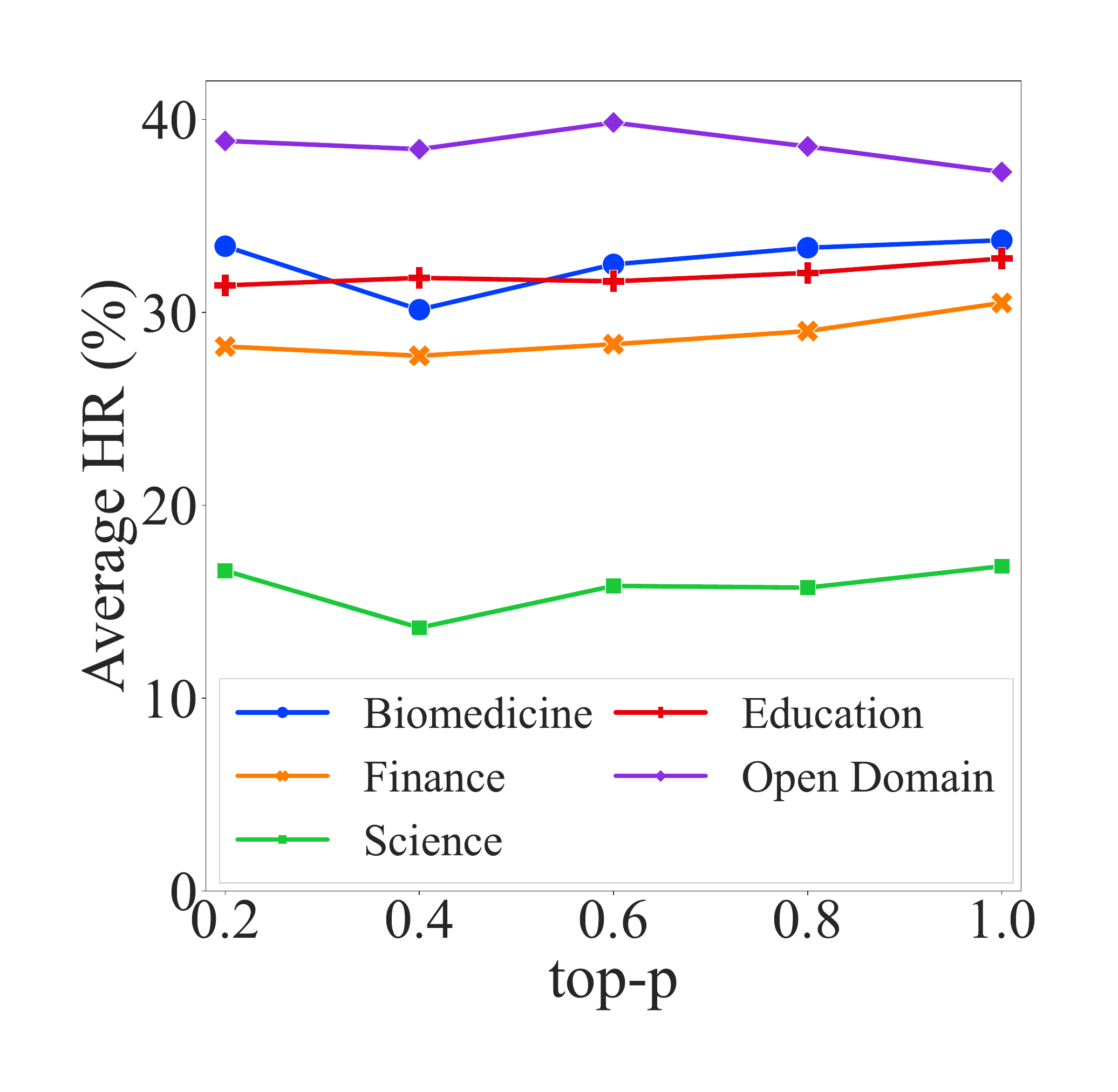}
	}
	\subfigure[Llama 2-Chat (7B)]{
		\centering
		\includegraphics[width=0.22\textwidth]{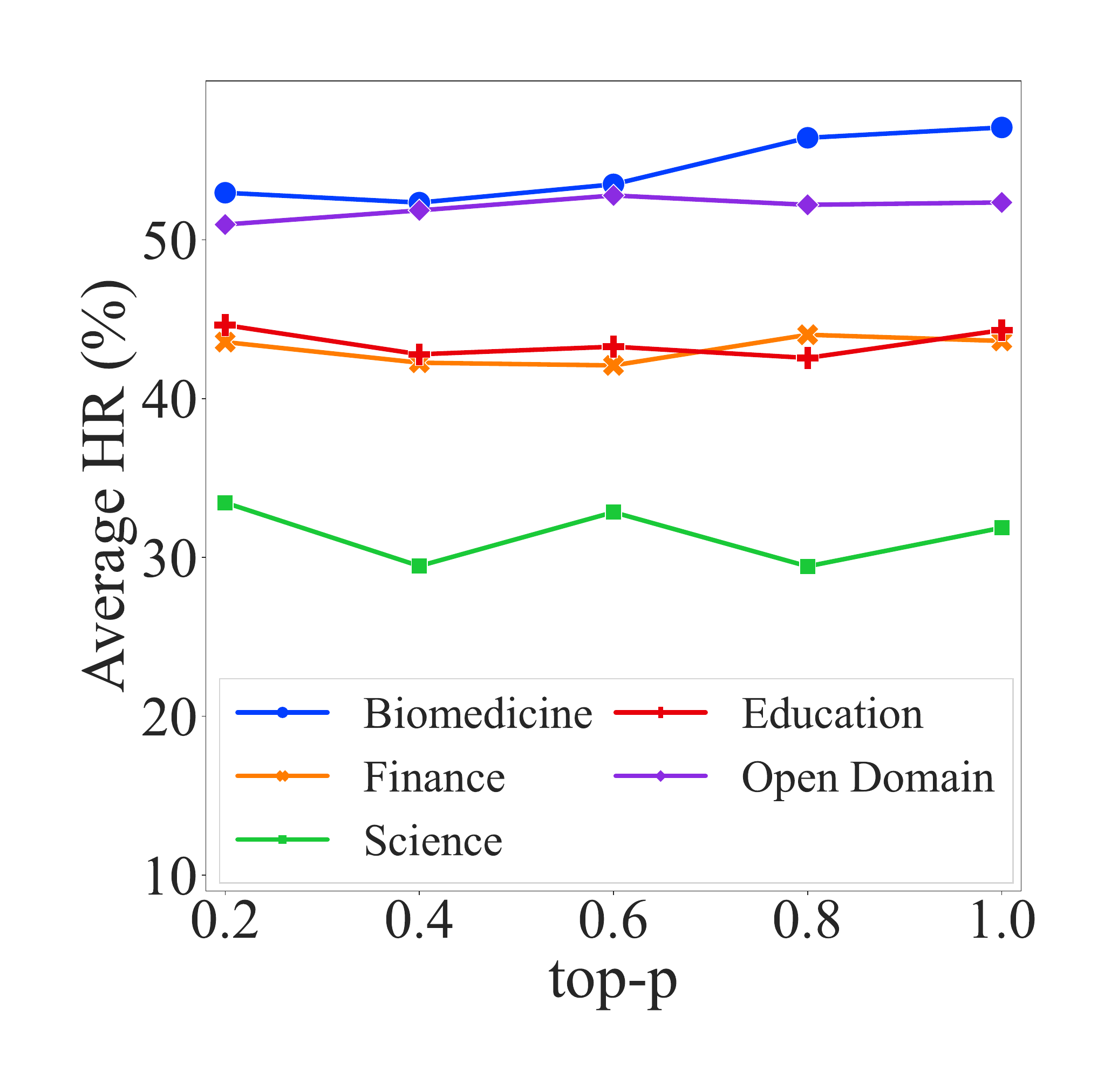}
	}
	\centering
	\caption{Average hallucination rate (\%) with varying $p$ in top-$p$ sampling.}
	\label{fig:vary_p}
	\vspace{-0.2cm}
\end{figure}

\paratitle{Decoding Strategies.} We explore the influence of four decoding strategies on LLM hallucinations including greedy search, top-$k$ sampling, top-$p$ sampling, and beam search. Specifically, we select an open-source Llama 2-Chat (7B) and closed-source ChatGPT, and set $k=20$, $p=0.5$, and the number of beams to $5$. For ChatGPT, due to the API constraint, we only investigate greedy search and top-$p$ sampling strategies. The results of different decoding strategies are shown in Table~\ref{tab:decoding_strategies}. We observe that diversity-enhanced decoding strategies such as top-$p$ sampling contribute to inducing more hallucinations in professional domains (\eg science), while greedy search will exacerbate the generation of hallucinations {in open-ended domains (\eg education). This phenomenon is even more pronounced in smaller models. Beam search can effectively balance the trade-off between the generation diversity and factuality.} 
Furthermore, to examine the correlation between diversity and hallucinations, we vary $p$ of top-$p$ sampling in the set $\{0.2, 0.4, 0.6, 0.8, 1.0\}$. We show the trend of average hallucination rate with varying $p$ in ChatGPT and Llama 2-Chat (7B) in Figure~\ref{fig:vary_p}. As can be seen, the hallucination rates of ChatGPT and Llama 2-Chat are highly sensitive to the varying of top-$p$ values within the professional domains, oscillating in biomedicine and science, while exhibiting minimal fluctuations in other open-ended domains.

\begin{table*}[t!]
\small
    \centering
    \begin{tabular}{c|l|p{8.5cm}}
    \toprule
       \rowcolor[gray]{.92} \multicolumn{1}{c}{\textbf{Domain}} & \multicolumn{1}{c}{\textbf{Question}} & \multicolumn{1}{c}{\textbf{ChatGPT Response}} \\
    \midrule
        \multirow{10}{*}{\textbf{Biomedicine}} & \multirow{10.5}{*}{\begin{tabular}[x]{@{}l@{}} Is the regulation of Vsr \\endonuclease independent of \\the growth phase of bacteria?  \end{tabular}} & \textcolor{tOrange}{During the exponential growth phase, when bacteria are actively dividing and replicating their DNA, DNA replication errors and mismatched bases can \textbf{more}} frequently occur. These errors can lead to mutations in the genetic code, introducing variations in the bacterial population. \textcolor{red}{\CheckmarkBold}  \\
        \cline{3-3}
        & & \textcolor{tOrange}{During the exponential growth phase, when bacteria are actively dividing and replicating their DNA, DNA replication errors and mismatched bases can \textbf{less}} frequently lead to significant consequences. This is because the rapid pace of cell division during the exponential growth phase may outpace the usual repair mechanisms that cells employ to correct such errors.\textcolor{red}{\XSolidBrush} \\
        \midrule
        \multirow{8}{*}{\textbf{Finance}} & \multirow{8.5}{*}{\begin{tabular}[x]{@{}l@{}} Record retention requirements for \\individuals in the U.S.?  \end{tabular}} & \textcolor{tOrange}{Generally, it is suggested to keep tax-related records for \textbf{at least}} seven years. This duration allows individuals and businesses to comply with various tax regulations and be prepared for any potential audits or inquiries from tax authorities.\textcolor{red}{\CheckmarkBold} \\
        \cline{3-3}
        &  & \textcolor{tOrange}{Generally, it is suggested to keep tax-related records for \textbf{at most}} seven years. This timeframe allows individuals and businesses to comply with the typical statute of limitations for tax audits and inquiries.\textcolor{red}{\XSolidBrush}  \\
    \midrule
      \multirow{6}{*}{\textbf{Science}} & \multirow{6.5}{*}{\begin{tabular}[x]{@{}l@{}} Do you agree with the claim that \\men are more susceptible to death \\due to pneumonia when compared \\to women?  \end{tabular}} & \textcolor{tOrange}{The gender gap in pneumonia mortality becomes more prominent with age. In older age groups, \textbf{men}} have a higher risk of dying from pneumonia compared to women.\textcolor{red}{\CheckmarkBold} \\ 
    \cline{3-3}
      &  & \textcolor{tOrange}{The gender gap in pneumonia mortality becomes more prominent with age. In older age groups, \textbf{women}} exhibit a higher susceptibility to severe pneumonia-related outcomes compared to men.\textcolor{red}{\XSolidBrush} \\
      \midrule
      \multirow{4}{*}{\textbf{Education}} & \multirow{4.5}{*}{\begin{tabular}[x]{@{}l@{}} what innovation did the roman \\numerals use ? \end{tabular}}  & \textcolor{tOrange}{The Roman numeral system introduced subtractive notation, where a \textbf{smaller}} value is placed before a larger value to indicate subtraction.\textcolor{red}{\CheckmarkBold}  \\ 
     \cline{3-3}
      &  & \textcolor{tOrange}{The Roman numeral system introduced subtractive notation, where a \textbf{larger}} value is placed before a smaller one to indicate subtraction.\textcolor{red}{\XSolidBrush}  \\
      \midrule
      \multirow{4}{*}{\textbf{Open Domain}} & \multirow{4.5}{*}{\begin{tabular}[x]{@{}l@{}} Can you give some details \\about Pluto? \end{tabular}} & \textcolor{tOrange}{Pluto is the \textbf{second smallest}} dwarf planet in our solar system after Eris.\textcolor{red}{\CheckmarkBold} \\ 
      \cline{3-3}
      &  &  \textcolor{tOrange}{Pluto is the \textbf{smallest}} and farthest known dwarf planet in our solar system, discovered by astronomer Clyde Tombaugh in 1930.\textcolor{red}{\XSolidBrush} \\
    \bottomrule
    \end{tabular}
    \caption{Illustrative examples for token-by-token generation showing that the LLM will commit to the previously generated token even if it might lead to hallucinations. The \textcolor{tOrange}{brown} span denotes the same generated prefix by ChatGPT, and the \textcolor{tOrange}{\textbf{bold}} font denotes the replaced token.}
    \label{tab:token_by_token}
\end{table*}

\paratitle{Quantization.} To understand the impact of quantization on hallucinations, we quantize Llama 2-Chat (7B) and (13B) at three precision
levels: 4-bit, 8-bit and 16-bit. Specially, we employ the library \texttt{bitesandbytes}\footnote{https://github.com/TimDettmers/bitsandbytes} to quantize the original 16-bit models to 8/4 bits by specifying the commands \texttt{load\_in\_8bit} and \texttt{load\_in\_4bit}, which focused on the quantization of {weights for LLMs}. The hallucination results of  quantized models are shown in Table~\ref{tab:quantization}. Despite with reduced memory footprint and accelerated inference rate, the 8-bit and 4-bit quantization results in an overall higher level of hallucination compared to the original 16-bit model. 
{In most cases, 8-bit quantization has a minimal impact on the hallucination of the model, while 4-bit quantization significantly increases the hallucination in the model's responses. In certain domains such as biomedicine, the hallucination rate of the quantized model will noticeably increases compared to the original model.} 

\paratitle{Token-by-Token Generation.} Most LLMs adopt the token-by-token manner to generate a token at a time. Once the generated part contains erroneous or unreasonable content, the model is difficult to complete the sentence correctly~\citep{Azaria-2023-arxiv-The}. To explore the drawback of token-by-token generation, we conduct a case study to qualitatively analyze how the generation paradigm leads to hallucinations. We present several examples across five domains in Table~\ref{tab:token_by_token}.
For each sample, we employ ChatGPT to generate two responses conditioned on two similar prefixes with the only difference being the last word. We can clearly observe that the LLM over-commits the errors in previous tokens and continuously complete the sentence incorrectly. For example, when we replace the degree words (\eg ``\emph{more}'' $\rightarrow$ ``\emph{less}''), ChatGPT does not identify the mistakes and keep generating the same sequence, leading to hallucinations. Besides, for the educational question about the roman numerals, when beginning with a prefix word ``\emph{larger}'', ChatGPT is unable to indicate subtraction by generating the right word ``\emph{after}'' (instead of ``\emph{before}''). 
Similarly, in the scientific domain, the real-world fact is that men have a higher risk of dying from pneumonia than woman. When we replace the prefix word ``\emph{men}'' with ``\emph{woman}'', ChatGPT is not able to complete the fact correctly by exchanging another expression.
In open domain, when we delete ``\emph{second}'' from the first prefix, ChatGPT incorrectly completes the sentence. In fact, one correct completion for the new prefix is ``\emph{Pluto is the smallest celestial body in the solar system that has ever been classified as a planet}''.

\begin{tcolorbox}
[colback=red!5!white,colframe=red!55!black,title=\textbf{\footnotesize \textsc{\ul{Effects of Inference Methods on llm hallucinations}}:}]
\begin{itemize}
[leftmargin=1mm]
\setlength\itemsep{0em}
    \item[\ding{224}] {\footnotesize 
    {\fontfamily{phv}\fontsize{8}{9}\selectfont
   Diversity-oriented decoding methods contribute to inducing more hallucinations in professional domains, while greedy search exacerbates hallucinations in open-ended domains. Beam search effectively balances the trade-off between diversity and factuality.}} 

     \item[\ding{224}] {\footnotesize 
    {\fontfamily{phv}\fontsize{8}{9}\selectfont
     In most cases, 8-bit quantization has a minimal impact on the model hallucinations, while 4-bit quantization significantly increases the hallucinations. The quantization method overall exacerbates the hallucinations in the responses of LLMs.
    }}

    \item[\ding{224}] {\footnotesize 
    {\fontfamily{phv}\fontsize{8}{9}\selectfont
   The token-by-token generation manner makes LLMs over-commit to the generation mistakes in previous tokens, which may be difficult to be correctly completed and leads to hallucinations.}} 
\end{itemize}
\end{tcolorbox}

%% file: sec/mitigation.tex
\section{Hallucination Mitigation}
Above, we have extensively discussed the possible sources for LLM hallucinations. To alleviate  the hallucinating behaviors of  LLMs, a number of hallucination mitigation studies have been proposed in existing literature. Specially, we invesitiage the effectiveness of several widely used methods, including reinforcement learning from human feedback (RLHF), retrieval augmentation, self-reflexion, advanced decoding, and prompt improvement. 

\begin{table*}[t]
	\small
	\centering
	\begin{tabular}{l c c c c c c c c c c}
		\toprule
		\multicolumn{1}{l}{\multirow{2.5}{*}{\textbf{Models}}} & \multicolumn{2}{c}{\textbf{Biomedicine}} & \multicolumn{2}{c}{\textbf{Finance}} & \multicolumn{2}{c}{\textbf{Science}} & \multicolumn{2}{c}{\textbf{Education}} & \multicolumn{2}{c}{\textbf{Open Domain}} \\ 
		\cmidrule(r){2-3}\cmidrule(r){4-5}\cmidrule(r){6-7}\cmidrule(r){8-9}\cmidrule(r){10-11}
		& MaHR  & MiHR & MaHR  & MiHR & MaHR  & MiHR & MaHR  & MiHR & MaHR  & MiHR \\ 
            \midrule[0.5pt]
            \textbf{Alpaca 7B} & 75.56 & 30.92 & 73.40 & 29.01 & 47.95 & 13.15 & 78.84 & 28.86 & 65.34 & 29.03 \\
            w/ RLHF & 67.02 & 28.32 & 70.06 & 28.65 & 47.00 & 13.83 & 63.95 & 26.33 & 55.29 & 25.02 \\
            \midrule[0.5pt]
            \textbf{Vicuna 7B} & 72.59 & 27.75 & 73.06 & 25.28 & 49.49 & 13.79 & 70.78 & 27.62 & 64.52 & 30.31 \\
            w/ RLHF & 70.85 & 25.90 & 70.33 & 23.87 & 48.50 & 13.45 & 68.32 & 25.39 & 53.75 & 25.78 \\
            \bottomrule
	\end{tabular}
	\caption{Evaluation results for Alpaca (7B) and Vicuna (7B) after RLHF alignment. }
        \vspace{-0.1cm}
	\label{tab:rlhf}
\end{table*}

\subsection{RLHF}

Reinforcement learning from human feedback (RLHF) is the process of fine-tuning language models with human feedback data to align with human values~\citep{Ouyang-arxiv-2022-Training}. Specially, ``3H alignment'' (\ie \emph{helpful}, \emph{honest}, and \emph{harmless}) is widely used as the alignment criterion in RLHF, where \emph{honest} is partially related to hallucination mitigation in LLM's  responses. Typically, employing RLHF to mitigate LLM hallucinations involves two steps: (1) collect hallucinated and non-hallucinated (human-preferred) responses to train a reward model; (2) fine-tune the LLM with the reward model's feedback using RL algorithms. 

\paratitle{Experimental Details.} We include the input questions in HaluEval 2.0 (besides the selected 1,000 test samples) as the initial prompt set. Following existing work~\citep{OpenAI-OpenAI-2023-GPT-4}, we first pass a prompt to an unaligned LLM (\eg Alpaca 7B) and get a response, then feed the prompt and response through GPT-4 to list all hallucinations. If there are any hallucinations, we iteratively employ GPT-4 to rewrite the response with up to 5 rounds until there are no any hallucinations. We keep \emph{(prompt, hallucinated response, refined response)} as comparison data to train a reward model. The reward model is usually initialized from the same model checkpoint as the LLM to be aligned. This practice ensures that the reward model ``knows'' what the unaligned LLM knows and prevents an information mismatch between them~\citep{Touvron-2023-llama2-arxiv}. Finally, we fine-tune the unaligned LLM with the reward model using PPO. Specifically, we apply this approach to two unaligned LLMs, \ie Alpaca (7B) and Vicuna (7B), and evaluate RLHF-aligned models on the selected 1,000 test samples.

\paratitle{Results and Analysis.}
The evaluation results are shown in Table~\ref{tab:rlhf}. We can clearly see that, after the RLHF alignment fine-tuning, the phenomenon of LLM hallucinations has been significantly alleviated in certain domains such as open domain. Through the RLHF process, the model generates more accurate facts and its responses become more concise without much irrelevant content, leading to less factuality hallucinations. Moreover, we can see that the effect of RLHF in mitigating hallucinations is dependent on the domains, exhibiting more pronounced hallucination reduction in biomedicine and open domain, while showing mild effectiveness in some highly professional domains such as science. In existing research, RLHF is mostly focused on open-ended domains but ignoring the alignment on specific domains for LLMs. We suggest that to make LLMs more versatile it is crucial to execute RLHF fine-tuning in broader domains.

\begin{tcolorbox}
[colback=red!5!white,colframe=red!55!black,title=\textbf{\footnotesize \textsc{\ul{Effects of RLHF on mitigating llm hallucinations}}:}]
\begin{itemize}
[leftmargin=1mm]
\setlength\itemsep{0em}
    
    \item[\ding{224}] {\footnotesize 
    {\fontfamily{phv}\fontsize{8}{9}\selectfont
   The RLHF fine-tuning can be helpful to alleviate the phenomenon of hallucinations in LLMs, but such effectiveness relies on domains, \ie exhibiting mild effects on highly professional domains such as science.}}

   \item[\ding{224}] {\footnotesize 
    {\fontfamily{phv}\fontsize{8}{9}\selectfont
   Future direction could involve performing RLHF fine-tuning in general and specific domains to make LLMs versatile, leading to improved robustness, generalization, and accuracy in fact-intensive tasks.}}

\end{itemize}
\end{tcolorbox}

\begin{table*}[t]
	\small
	\centering
	\begin{tabular}{l c c c c c c c c c c}
		\toprule
		\multicolumn{1}{l}{\multirow{2.5}{*}{\textbf{Models}}} & \multicolumn{2}{c}{\textbf{Biomedicine}} & \multicolumn{2}{c}{\textbf{Finance}} & \multicolumn{2}{c}{\textbf{Science}} & \multicolumn{2}{c}{\textbf{Education}} & \multicolumn{2}{c}{\textbf{Open Domain}} \\ 
		\cmidrule(r){2-3}\cmidrule(r){4-5}\cmidrule(r){6-7}\cmidrule(r){8-9}\cmidrule(r){10-11}
		& MaHR  & MiHR & MaHR  & MiHR & MaHR  & MiHR & MaHR  & MiHR & MaHR  & MiHR \\ 
            \midrule[0.5pt]
            \textbf{ChatGPT} & 48.75 & 14.03 & 46.84 & 13.55 & 24.14 & 6.39 & 53.44 & 17.19 & 59.77 & 17.93 \\
            w/ Retrieval & 23.98 & 12.18 & 38.85 & 15.25 & 21.15 & 6.19 & 35.97 & 15.52 & 34.15 & 16.51 \\
            \midrule[0.5pt]
            \textbf{Llama 2-Chat 7B} & 69.12 & 26.69 & 69.41 & 24.59 & 49.25 & 14.05 & 71.52 & 27.74 & 77.35 & 33.15 \\
            w/ Retrieval & 45.13 & 14.67 & 63.92 & 21.25 & 34.81 & 10.02 & 62.84 & 25.03 & 55.81 & 24.41 \\
            \midrule[0.5pt]
            \textbf{Llama 2-Chat 13B} & 70.56 & 26.63 & 69.95 & 23.85 & 42.21 & 13.02 & 69.14 & 26.71 & 76.34 & 32.48 \\
            w/ Retrieval & 43.62 & 14.00 & 64.74 & 21.19 & 32.65 & 9.76 & 62.59 & 23.73 & 47.62 & 23.06 \\
            \midrule[0.5pt]
            \textbf{Vicuna 7B} & 72.59 & 27.75 & 73.06 & 25.28 & 49.49 & 13.79 & 70.78 & 27.62 & 64.52 & 30.31 \\
            w/ Retrieval & 43.24 & 13.80 & 61.41 & 21.49 & 34.75 & 9.35 & 60.28 & 21.54 & 51.15 & 21.37 \\
            \midrule[0.5pt]
            \textbf{Vicuna 13B} & 71.36 & 27.76 & 67.86 & 22.75 & 47.45 & 13.09 & 68.07 & 27.53 & 74.31 & 33.39 \\
            w/ Retrieval & 43.20 & 13.39 & 55.87 & 18.24 & 32.79 & 10.05 & 60.94 & 24.57 & 52.27 & 21.38 \\
            \bottomrule
	\end{tabular}
	\caption{Evaluation results for LLMs with retrieval augmentation. }
        \vspace{-0.1cm}
	\label{tab:retrieval}
\end{table*}

\subsection{Retrieval Augmentation}

Retrieval augmentation has been widely used to enhance the capacities of LLMs~\citep{ren2023investigating} and is generally considered as one of the effective approaches to alleviate hallucination~\citep{Li-arxiv-2023-HaluEval, ji2023survey}.
The basic idea of retrieval-augmented hallucination mitigation is to first retrieve a small set of supporting documents from a large-scale document corpus (\eg Wikipedia) based on a user query and then the LLM can generate an accurate answer to the question conditioned on the retrieved documents.


\paratitle{Experimental Details.}
In our experiments, we conduct web search by retrieving documents from the whole web. Specially, we utilize the input question verbatim as query and request a call to Bing Search via API\footnote{https://www.microsoft.com/en-us/bing/apis/bing-web-search-api}. For the search results, we only use the snippets of webpages as the retrieved documents.
First, to examine the impact of the number of retrieved documents on hallucination mitigation, we use the top-$k$ documents as context and vary $k$ in the set $\{1, 2, 5, 10\}$.
Second, to further validate the effect of the relevance of retrieved documents on hallucination mitigation, we randomly sample one document from top-$k$ documents. Here, the variance of $k$ reflects four levels of relevance to the question, ranging from strong to weak relevance.  

\begin{figure}[t]
	\centering
	\subfigure[ChatGPT]{
		\centering
		\includegraphics[width=0.22\textwidth]{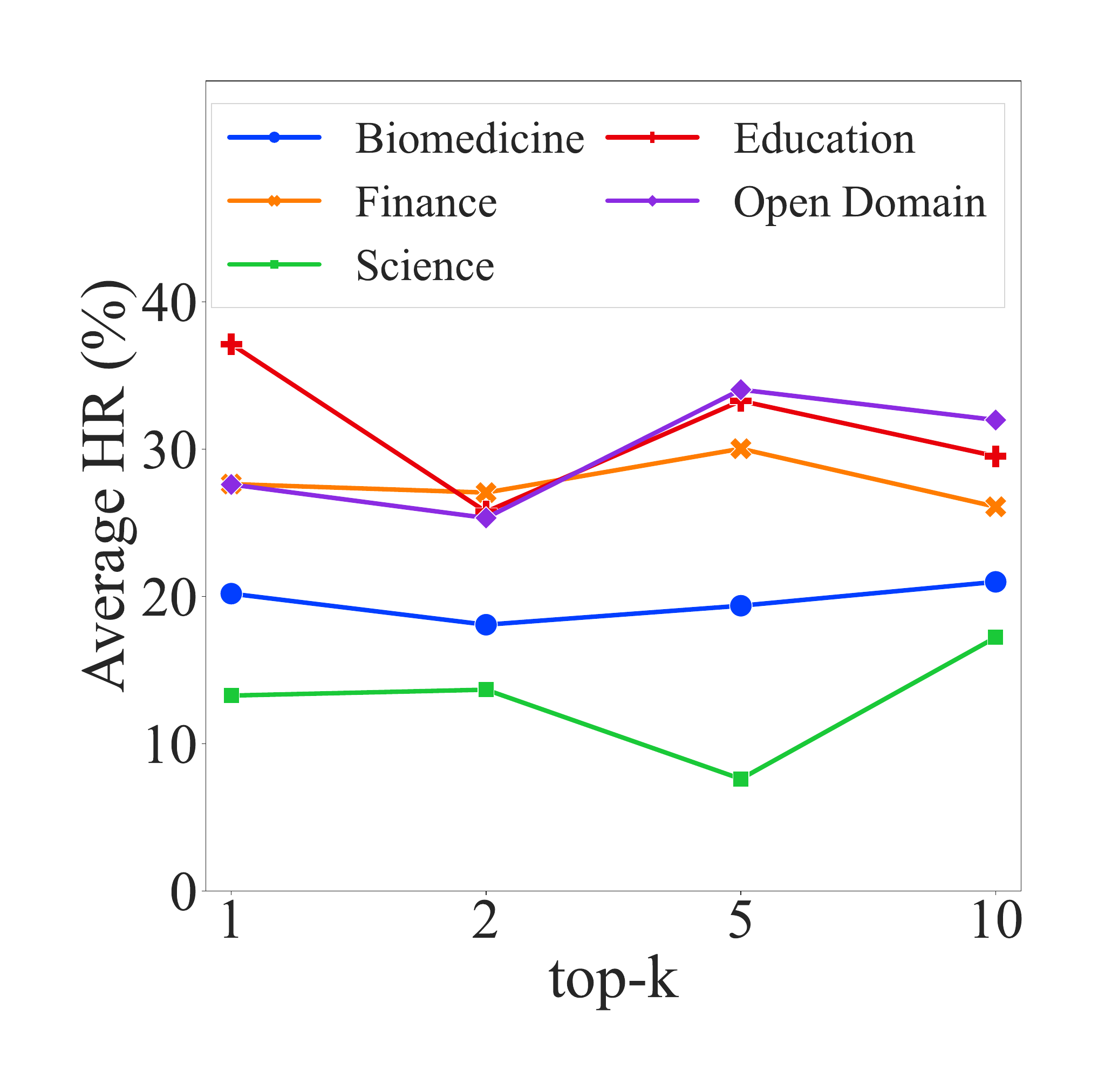}
	}
	\subfigure[Llama 2-Chat (7B)]{
		\centering
		\includegraphics[width=0.22\textwidth]{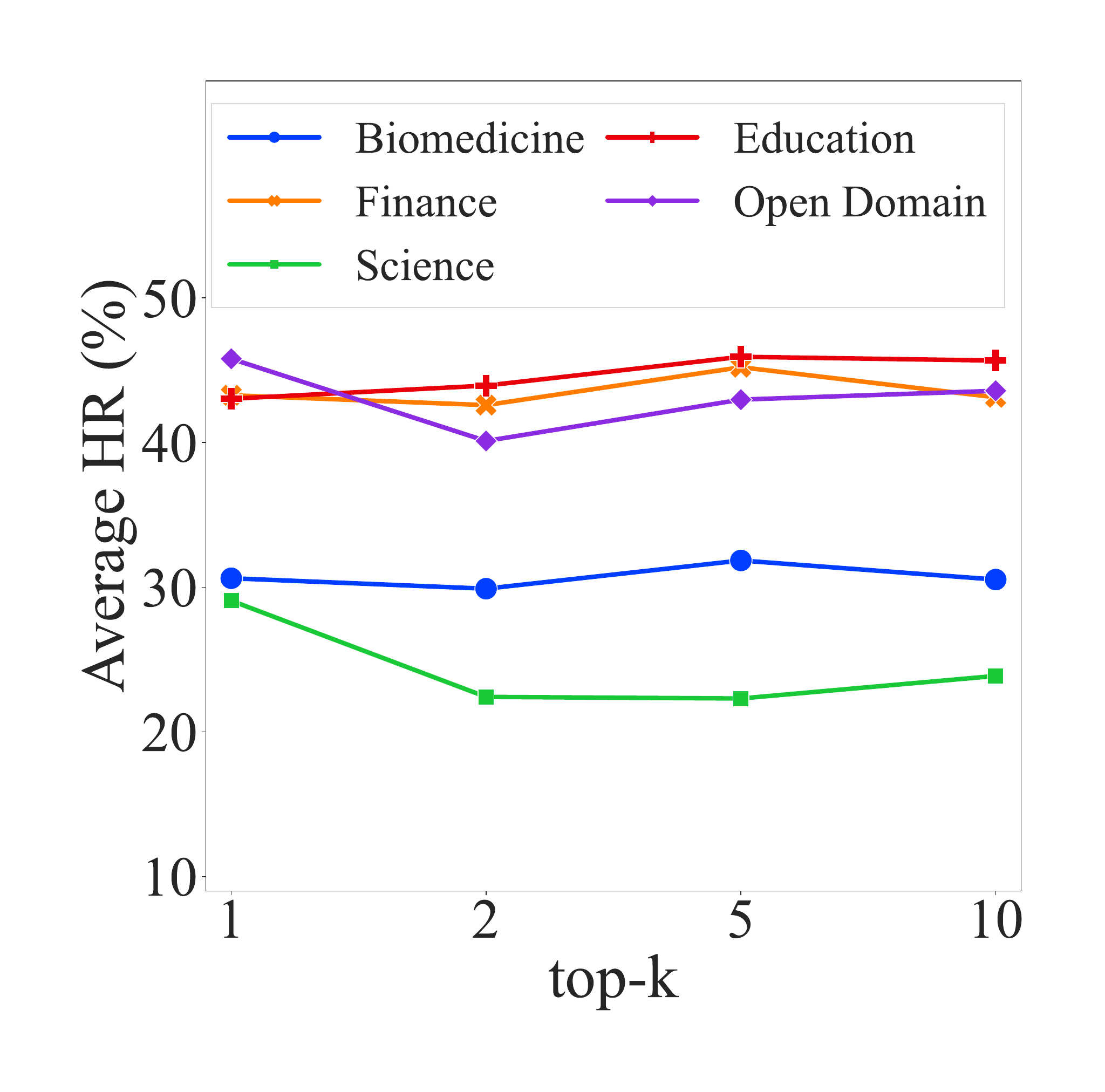}
	}
	\centering
	\caption{Average hallucination rate (\%) of using top-$k$ retrieved documents as context.}
	\label{fig:vary_k_all}
	\vspace{-0.2cm}
\end{figure}

\begin{figure}[t]
	\centering
	\subfigure[ChatGPT]{
		\centering
		\includegraphics[width=0.22\textwidth]{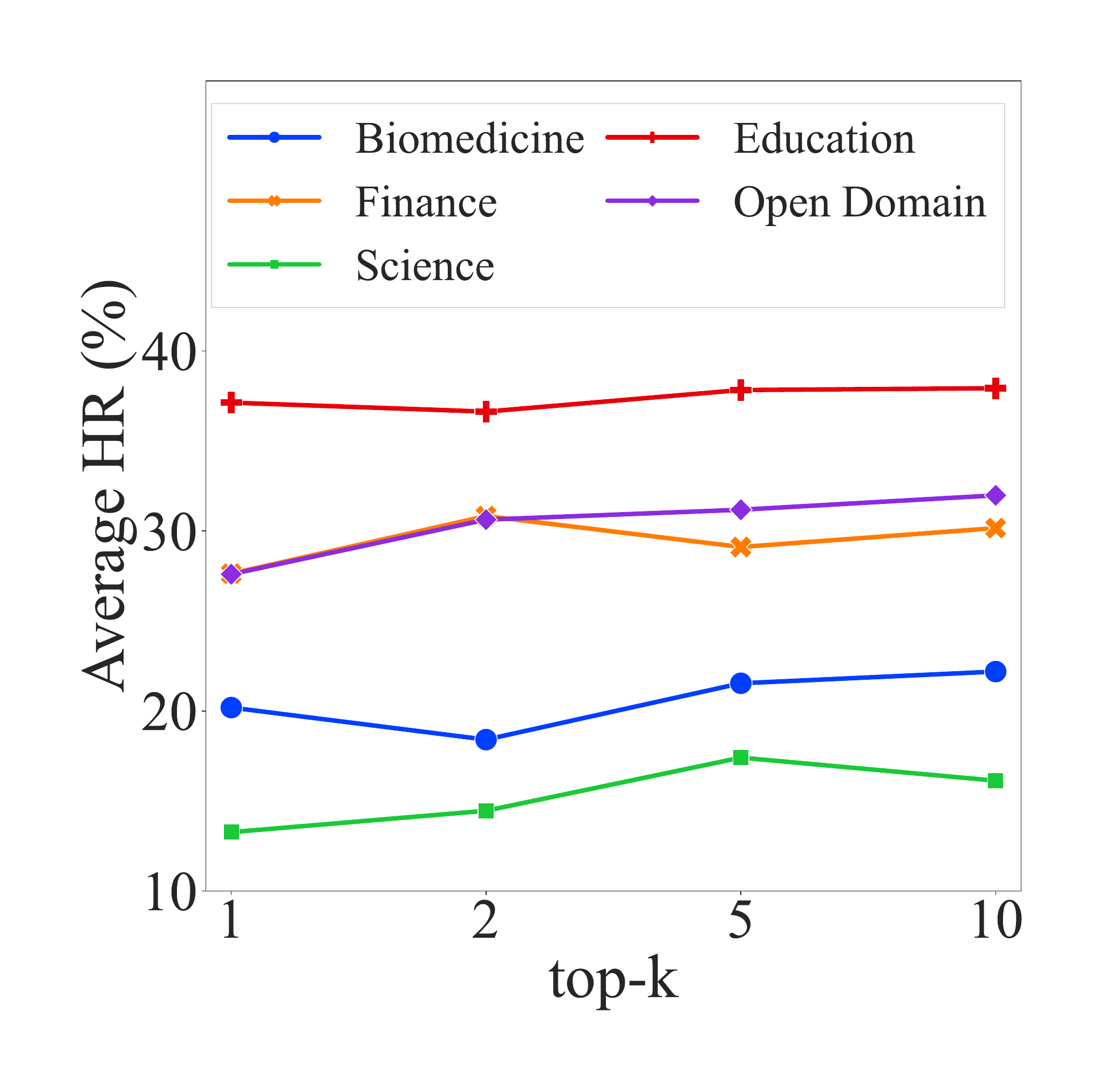}
	}
	\subfigure[Llama 2-Chat (7B)]{
		\centering
		\includegraphics[width=0.22\textwidth]{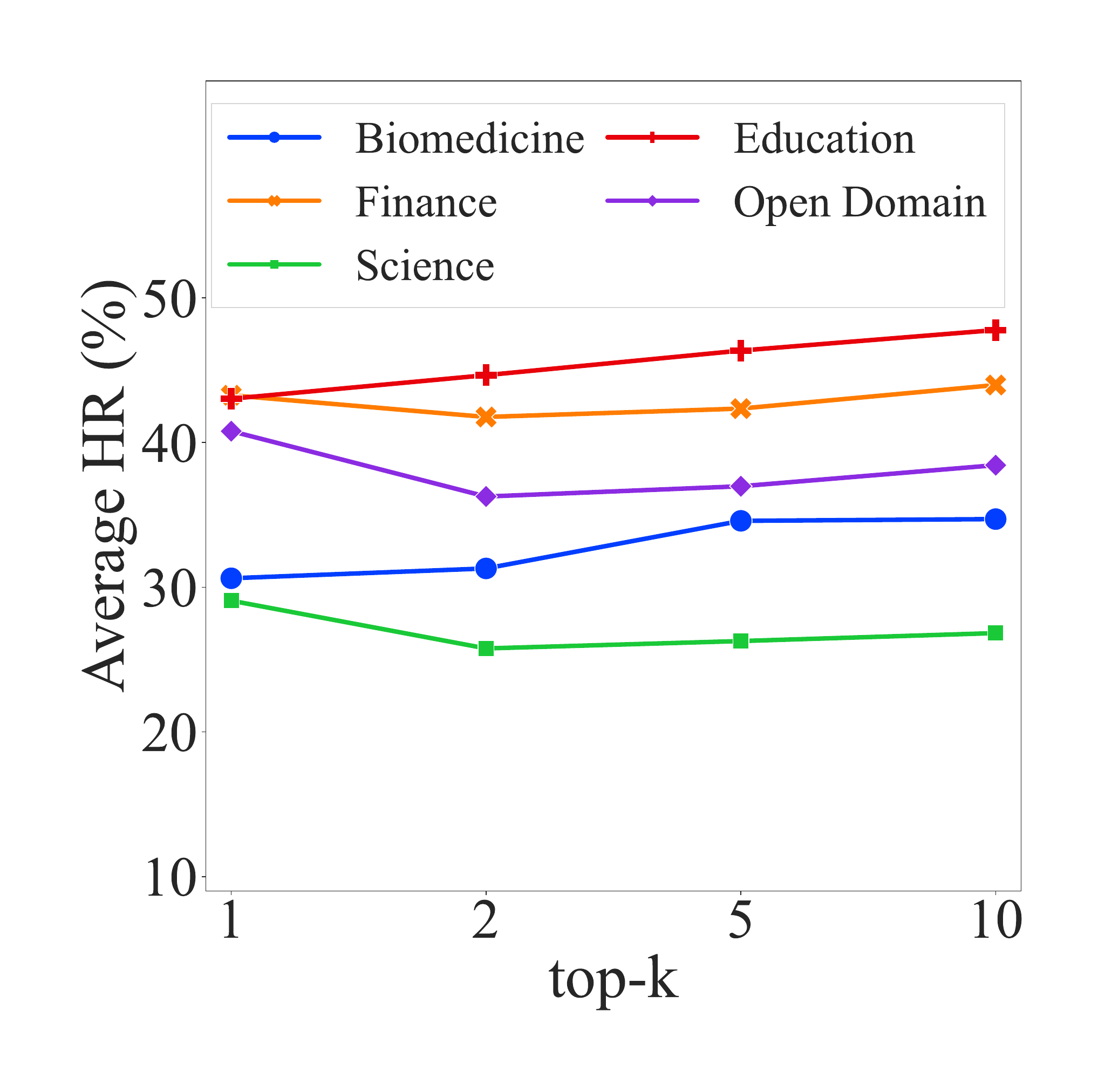}
	}
	\centering
	\caption{Average hallucination rate (\%) of sampling one document from top-$k$ retrieved documents.}
	\label{fig:vary_k_one}
	\vspace{-0.2cm}
\end{figure}

\paratitle{Results and Analysis.}
In Figure~\ref{fig:vary_k_all}, we show the results of using top-$k$ documents as evidence. 
We can clearly see that ChatGPT and Llama 2-Chat mostly produce less hallucinations in top-$2$ retrieval, except that ChatGPT prefers top-$5$ in science. This is because that including more documents as the context of LLMs can bring more noises into the generation process, leading to higher level of hallucination. 
Therefore, we conduct top-$2$ retrieval for several LLMs and present the results in Table~\ref{tab:retrieval}. It can be verified that retrieval can significantly mitigate the hallucinations in the responses of LLMs. The effectiveness of retrieval is more pronounced in smaller models (\eg Llama 2-Chat 7B), as the hallucination rate of larger models (\eg ChatGPT) has already been relatively low and smaller models acquire limited world knowledge.
Besides, we present the effect of the relevance between question and document by randomly selecting one document from top-$k$ results. The results in Figure~\ref{fig:vary_k_one} shows that the lower the relevance between the retrieved document and the question, the more likely the model is to generate hallucinations. However, more capable models like ChatGPT are less sensitive to the relevance of retrieved documents.

\begin{tcolorbox}
[colback=red!5!white,colframe=red!55!black,title=\textbf{\footnotesize \textsc{\ul{Effects of Retrieval on mitigating llm hallucinations}}:}]
\begin{itemize}
[leftmargin=1mm]
\setlength\itemsep{0em}
    
    \item[\ding{224}] {\footnotesize 
    {\fontfamily{phv}\fontsize{8}{9}\selectfont
   Retrieval can significantly mitigate LLM hallucinations. This effectiveness is more pronounced for smaller models (\eg Llama 2-Chat 7B), as smaller models only acquire limited world knowledge and larger models (\eg ChatGPT) show a low level of hallucinations.}}

   \item[\ding{224}] {\footnotesize 
    {\fontfamily{phv}\fontsize{8}{9}\selectfont
   The relevance between question and document highly affects the effectiveness of retrieval. The lower the relevance between the retrieved document and question, the more likely the model is to generate hallucinations. However, more capable models like ChatGPT are less sensitive to the variance of relevance.}}

\end{itemize}
\end{tcolorbox}

\begin{figure*}[t]
    \centering
    \includegraphics[width=\textwidth]{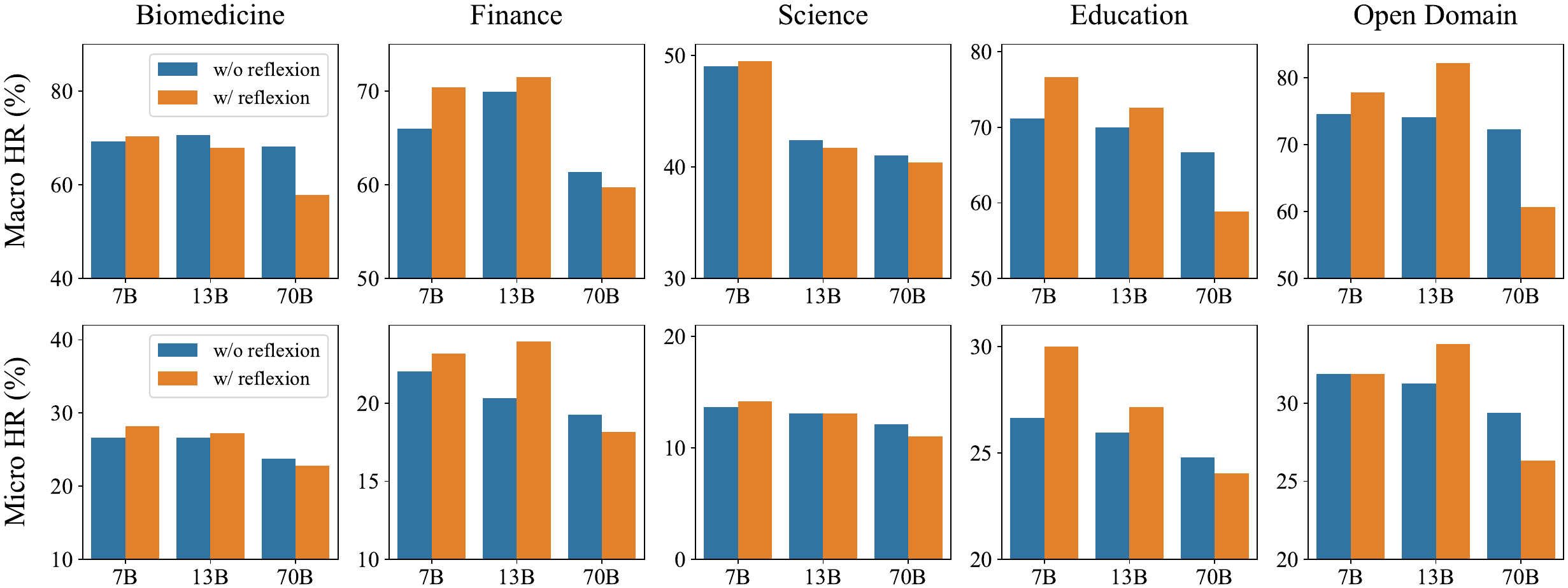}
    \caption{Macro and micro hallucination rate (\%) for Llama 2-Chat (7B, 13B and 70B) with or without self-reflexion.
    }
    \label{fig:reflexion}
\end{figure*}

\subsection{Self-Reflexion}

{Reflexion}~\citep{shinn2023reflexion} is an effective means through which LLMs can learn from and rectify their mistakes. 
Specially, the prior failure trials are transformed as textual feedback, which would be incorporated  as additional 
context for the LLM itself. With the guidance of feedback, the LLM can generate an improved plan for the next attempt, called \emph{self-reflexion}. 
Existing research~\citep{Ji2023towards, Dhuliawala-2023-CoRR-Chain} has demonstrated that self-reflexion is an effective method for hallucination mitigation.
However, self-reflexion is a complex and advanced ability that requires mistake perception, feedback summarization, and behavioral planning. 
In light of this, we aim to explore to which extent self-reflexion can mitigate hallucinations and how the capacities of LLMs affect the reflexion performance on hallucination mitigation. 


\paratitle{Experimental Details.}
We conduct self-reflexion experiments on Llama 2-Chat 7B, 13B, and 70B, to examine its effect on hallucination mitigation. 
Specifically, for a given query, we first obtain the initial model response, then require the model itself with instructions to reflect on whether its response contains any errors. If errors are detected, the model is prompted to provide a corrected response. This process is repeated until the model response is error-free or the maximum number of reflection iterations is reached. We set the maximum number of reflection iterations to $5$. In addition, we also test these models without self-reflexion for performance comparison.

\paratitle{Results and Analysis.}
Figure~\ref{fig:reflexion} illustrates the relationship between the model's scale and its effectiveness in mitigating hallucinations through self-reflexion. We can clearly see that only when reaching a certain scale (\ie 70B in our experiments), the LLM possesses the self-reflective ability to mitigate hallucinations in its responses. For Llama 2-Chat 7B and 13B, self-reflexion, on the contrary, can even result in more erroneous responses. 
We found that due to the limited capacity of smaller models, the reflective behavior instead makes them suspect their original correct answers and generates wrong ones. 
Furthermore, the benefits obtained from self-reflexion are domain-sensitive, \eg self-reflexion reduces the LLM hallucinations slightly in the fields of finance and science, while significantly in the open domain. 

\begin{tcolorbox}
[colback=red!5!white,colframe=red!55!black,title=\textbf{\footnotesize \textsc{\ul{Effects of Self-Reflexion on mitigating llm hallucinations}}:}]
\begin{itemize}
[leftmargin=1mm]
\setlength\itemsep{0em}
    
    \item[\ding{224}] {\footnotesize 
    {\fontfamily{phv}\fontsize{8}{9}\selectfont
   Self-reflexion is a kind of emergent ability when the scale of LLMs achieves a certain level (\eg 70B). For smaller models, self-reflexion would make them suspect their original correct answers and mislead them to generate wrong ones with hallucinations.}}

     \item[\ding{224}] {\footnotesize 
    {\fontfamily{phv}\fontsize{8}{9}\selectfont
     The effectiveness of self-reflexion on hallucination mitigation is domain-sensitive, \ie slightly in finance and science, while significantly in open domain.
    }}
\end{itemize}
\end{tcolorbox}

\begin{table*}[t]
	\small
	\centering
	\begin{tabular}{l c c c c c c c c c c}
		\toprule
		\multicolumn{1}{l}{\multirow{2.5}{*}{\textbf{Models}}} & \multicolumn{2}{c}{\textbf{Biomedicine}} & \multicolumn{2}{c}{\textbf{Finance}} & \multicolumn{2}{c}{\textbf{Science}} & \multicolumn{2}{c}{\textbf{Education}} & \multicolumn{2}{c}{\textbf{Open Domain}} \\ 
		\cmidrule(r){2-3}\cmidrule(r){4-5}\cmidrule(r){6-7}\cmidrule(r){8-9}\cmidrule(r){10-11}
		& MaHR  & MiHR & MaHR  & MiHR & MaHR  & MiHR & MaHR  & MiHR & MaHR  & MiHR \\ 
            \midrule[0.5pt]
            \textbf{Llama 2-Chat 7B} &  \\
            w/ greedy search & 69.12 & 26.69 & 69.41 & 24.59 & 49.25 & 14.05 & 71.52 & 27.74 & 77.35 & 33.15 \\
            w/ top-$p$ sampling & 79.70 & 34.48 & 63.36 & 21.25 & 49.20 & 13.13 & 61.45 & 25.47 & 74.86 & 30.70 \\
            w/ greedy-nucleus & 69.54 & 25.75 & 72.63 & 25.45 & 47.50 & 14.25 & 73.26 & 28.76 & 83.05 & 36.16 \\
            w/ factual-nucleus & 75.00 & 29.14 & 70.68 & 24.52 & 41.00 & 11.25 & 68.39 & 26.55 & 76.67 & 32.17 \\
            \bottomrule
	\end{tabular}
	\caption{Evaluation results of Llama 2-Chat (7B) using our designed greedy-nucleus and factual-nucleus sampling.}
	\label{tab:advanced_decoding}
\end{table*}

\subsection{Advanced Decoding}

{According to the results and analysis in Section~\ref{sec:inference_method}, to reduce the hallucinations while keeping the text quality requires a trade-off between the generation diversity and factuality.}  Following prior work~\citep{LeePXPFSC22,Bradley-2023-arxiv-Quality,Shi-2023-arxiv-Trusting}, we design two simple yet effective decoding methods that can flexibly switch between greedy search and top-$p$ sampling, to balance the generation diversity and factuality. The first decoding strategy assumes that when the model has high confidence in predicting the next word, it should employ greedy search, otherwise it should utilize top-$p$ sampling, that we refer to as \emph{greedy-nucleus sampling}. The second strategy hypothesizes that the randomness of generation in the latter part of a sentence should be smaller than the beginning since there is no preceding text at the start of a sentence, which is called \emph{factual-nucleus sampling}~\citep{LeePXPFSC22}.

\paratitle{Experimental Details.} 
We detail the two advanced decoding methods as follows:

$\bullet$~\emph{Greedy-nucleus sampling}: Following previous studies~\citep{kadavath2022language,LiTZWNW23}, we use \emph{entropy} to quantify the confidence of the model. Thus, if the entropy $e$ of the word distribution is smaller than a threshold $\eta$ (high confidence), we adopt greedy search to generate the next word $w_t$, otherwise, we employ top-$p$ sampling to generate. We formulate this method as:
\begin{equation}
    w_t = \left\{
    \begin{aligned}
        &\arg\max P(w_t|w_{1,...,w_{t-1}}), \quad e \leq \eta \\
        &\sum P(w_t|w_{1,...,w_{t-1}})\geq p, \quad e > \eta 
    \end{aligned}
    \right.
\end{equation}

$\bullet$~\emph{Factual-nucleus sampling}: Following previous work~\citep{LeePXPFSC22}, the probability $p_t$ in top-$p$ sampling to generate the $t$-th token can be formulated as:
\begin{equation}
    p_t = \max \{\beta, \ p \times \lambda^{t-1}\}, 
\end{equation}
where $\lambda$ is the decay factor that is used to decay the probability $p$ at each step to reduce the randomness through time, and $\beta$ is set as a lower bound of $p$ to guarantee a certain degree of diversity. We reset $p$ to the default value at the beginning of generating a new sentence.
We apply the two decoding methods to Llama 2-Chat (7B) and compare to the original greedy search and top-$p$ sampling methods.

\paratitle{Results and Analysis.} We present the results of the two diversity-factuality balanced decoding methods in Table~\ref{tab:advanced_decoding}.
As can be observed that, our proposed greedy-nucleus sampling strategy achieves comparable or even better results than the original decoding strategies in the domains of biomedicine and science, while the factual-nucleus introduced in previous work~\citep{LeePXPFSC22} performs well in the science, education, and open domains. It can be concluded that balancing the generation diversity and factuality benefits the reduction of hallucinations and the retention of text quality. However, our designed sampling strategies are sensitive to the choices of hyper-parameters such as the confidence threshold and decay factor. Therefore, we leave designing more effective decoding approaches to mitigating hallucinations in future work.

\begin{tcolorbox}
[colback=red!5!white,colframe=red!55!black,title=\textbf{\footnotesize \textsc{\ul{Effects of Advanced Decoding on mitigating llm hallucinations}}:}]
\begin{itemize}
[leftmargin=1mm]
\setlength\itemsep{0em}
    
    \item[\ding{224}] {\footnotesize 
    {\fontfamily{phv}\fontsize{8}{9}\selectfont
   Balancing the diversity and factuality during the generation process benefits the reduction of hallucinations and the retention of text quality. }}

   \item[\ding{224}] {\footnotesize 
    {\fontfamily{phv}\fontsize{8}{9}\selectfont
   Future directions could involve designing more effective decoding approaches to mitigating hallucinations for LLMs, considering the characteristics of specific domains and the black-box property of LLMs.}}
\end{itemize}
\end{tcolorbox}

\begin{table*}[t]
	\small
	\centering
	\begin{tabular}{l c c c c c c c c c c}
		\toprule
		\multicolumn{1}{l}{\multirow{2.5}{*}{\textbf{Models}}} & \multicolumn{2}{c}{\textbf{Biomedicine}} & \multicolumn{2}{c}{\textbf{Finance}} & \multicolumn{2}{c}{\textbf{Science}} & \multicolumn{2}{c}{\textbf{Education}} & \multicolumn{2}{c}{\textbf{Open Domain}} \\ 
		\cmidrule(r){2-3}\cmidrule(r){4-5}\cmidrule(r){6-7}\cmidrule(r){8-9}\cmidrule(r){10-11}
		& MaHR  & MiHR & MaHR  & MiHR & MaHR  & MiHR & MaHR  & MiHR & MaHR  & MiHR \\ 
            \midrule[0.5pt]
            \textbf{ChatGPT} & \\
            w/ base prompt & 48.75 & 14.03 & 46.84 & 13.55 & 24.14 & 6.39 & 53.44 & 17.19 & 59.77 & 17.93 \\
            + {manual desc} & 45.64 & 13.91 & 39.20 & 11.18 & 22.34 & 5.28 & 55.68 & 17.73 & 64.52 & 20.31 \\
            + {manual demo} & 42.71 & 14.89 & 40.74 & 12.12 & 25.27 & 7.11 & 56.41 & 19.24 & 44.72 & 21.88 \\
            + {domain info} &  51.02 & 14.40 & 42.21 & 13.17 & 26.13 & 6.43 & 53.76 & 16.05 & 61.83 & 19.80 \\
            + {character role} & 52.50 & 14.67 & 44.67 & 12.72 & 26.13 & 6.33 & 54.30 & 17.11 & 60.85 & 19.37 \\
            + {zero-shot cot} & 46.60 & 13.51 & 41.33 & 12.35 & 24.62 & 6.26 & 53.11 & 15.96 & 56.90 & 17.21 \\
            + {few-shot cot} & 38.98 & 13.94 & 46.88 & 11.88 & 21.00 & 5.07 & 57.79 & 25.22 & 55.12 & 20.71 \\
            \midrule[0.5pt]
            \textbf{Llama 2-Chat 7B} &  \\
            w/ {base prompt} & 69.12 & 26.69 & 69.41 & 24.59 & 49.25 & 14.05 & 71.52 & 27.74 & 77.35 & 33.15 \\
            + {manual desc} & 68.02 & 26.46 & 74.36 & 25.01 & 42.50 & 12.10 & 76.16 & 30.97 & 79.39 & 33.23 \\
            + {manual demo} & 69.70 & 27.90 & 66.33 & 24.61 & 45.00 & 12.27 & 71.01 & 27.02 & 66.88 & 31.84 \\
            + {domain info} & 77.66 & 30.48 & 71.72 & 24.55 & 45.50 & 13.72 & 78.82 & 30.72 & 78.61 & 36.33 \\
            + {character role} & 73.23 & 32.12 & 73.87 & 25.60 & 47.50 & 13.42 & 73.21 & 27.67 & 84.62 & 35.79 \\
            + {zero-shot cot} & 77.84 & 30.34 & 78.61 & 27.03 & 50.25 & 15.99 & 79.61 & 30.77 & 69.12 & 31.20 \\
            + {few-shot cot} & 71.21 & 28.47 & 73.58 & 25.70 & 48.00 & 15.00 & 70.66 & 29.99 & 71.93 & 32.68 \\
            \bottomrule
	\end{tabular}
	\caption{Evaluation results of ChatGPT and Llama 2-Chat (7B) using different prompt improvement strategies. }
	\label{tab:prompt_improvement}
\end{table*}

\subsection{Prompt Improvement}
According to the results and analysis in Section~\ref{sec:prompt_design}, we improve the task description, question expression, and in-context demonstrations in the original prompt. Furthermore, chain-of-thought (CoT) has been proven to be helpful in hallucination mitigation~\citep{Wang-arxiv-2023-Knowledge}, so we explore adding CoT reasoning into our prompt.

\paratitle{Experimental Details.}
We improve the original base prompt from the following aspects:

$\bullet$~We incorporate \emph{Base prompt}, \emph{Manual description prompt}, and \emph{Manual in-context prompt} from Section~\ref{sec:prompt_design} for comparison.

$\bullet$~\emph{Domain info prompt}: injecting domain information into the task description in the base prompt.

$\bullet$~\emph{Character role prompt}: defining a particular role (\eg scientist) for the system in the base prompt.

$\bullet$~\emph{Zero-shot cot prompt}: adding zero-shot CoT to the base prompt by prepending ``\texttt{Let's think step-by-step}''.

$\bullet$~\emph{Few-shot cot prompt}: adding few-shot CoT to the base prompt by injecting CoT examples.

Similarly, we test ChatGPT and Llama 2-Chat (7B) with these improved prompts to generate responses. For \emph{few-shot cot prompt}, the few-shot reasoning examples are manually-written and different for each domain.

\paratitle{Results and Analysis.} We show the prompt improvement results in Table~\ref{tab:prompt_improvement}. We can found that injecting domain information into task description or defining a character role for the system has an oscillatory effect on mitigating the hallucinations in LLM's responses. In the domains of finance and science, the two prompt improvement strategies can help LLMs generate more accurate responses. For Chain-of-Thought (CoT) prompting, its effect on hallucination mitigation heavily depends on the specific LLMs. For larger models like ChatGPT which possess exceptional reasoning capabilities, zero-shot or few-shot CoT reasoning can significantly benefit reducing the hallucinations. While for smaller models like Llama 2-Chat (7B) with limited reasoning abilities, leveraging CoT reasoning fails to effectively eliminate hallucinations and instead exacerbates the presence of hallucinations in their responses.
Therefore, when engaging in prompt engineering, it is crucial to comprehensively consider factors such as model size, domain characteristics, and task difficulty.

\begin{tcolorbox}
[colback=red!5!white,colframe=red!55!black,title=\textbf{\footnotesize \textsc{\ul{Effects of Prompt Improvement on mitigating llm hallucinations}}:}]
\begin{itemize}
[leftmargin=1mm]
\setlength\itemsep{0em}
    
    \item[\ding{224}] {\footnotesize 
    {\fontfamily{phv}\fontsize{8}{9}\selectfont
   Injecting domain information into the task description or defining a character role for the AI system has an oscillatory effect on hallucination mitigation. }}

   \item[\ding{224}] {\footnotesize 
    {\fontfamily{phv}\fontsize{8}{9}\selectfont
   The impact of CoT reasoning on hallucination mitigation heavily depends on the specific LLMs with varied reasoning abilities, \ie benefiting larger models like ChatGPT while exacerbating hallucinations for smaller models like Llama 2-Chat (7B).}}

     \item[\ding{224}] {\footnotesize 
    {\fontfamily{phv}\fontsize{8}{9}\selectfont
     When engaging in prompt engineering, it is crucial to comprehensively consider factors such as model size, domain characteristics, and task difficulty.
    }}
\end{itemize}
\end{tcolorbox}

%% file: sec/related.tex
\section{Related Work}

Hallucination has been a fundamental challenge in LLMs, receiving extensive attention in existing literature~\citep{Huang-arxiv-2023-A,ji2023survey,Zhang-CoRR-2023-Siren,Li-arxiv-2023-HaluEval}.
We discuss the related study in two aspects, namely hallucination source/detection and mitigation. 

\paratitle{Hallucination Source and Detection\footnote{Since the understanding of the hallucination source can be employed to devise new detection methods, a number of studies often jointly discussed the two parts. Thus, we discuss the source and detection in this single part. }.} 
To understand and detect the hallucination in LLMs, several existing studies focus on utilizing the LLM itself to as the tool to study the hallucinated content. For LLMs with access to their internal states, we can delve into the inner workings of the model to explore the principles behind hallucinations~\citep{varshney2023stitch,yuksekgonul2023attention,azaria2023internal}. 
Typically, the internal states that can be studied by examining the output logit values, the hidden layer activations, and the attention states. 
For example, \citet{varshney2023stitch} leveraged the output logit values of the model as a signal of hallucinations to estimate the uncertainty of responses. \citet{azaria2023internal} employed the hidden layer activations of the model for determining the truthfulness of generated statements. \citet{yuksekgonul2023attention} identified factual errors by employing a straightforward probe on the LLM's attention towards constraint tokens. For  models that can only be accessed through API calls, hallucinations are typically studied by analyzing the relationship between input prompts and the model's output responses~\citep{rawte2023exploring,manakul2023selfcheckgpt,yao2023llm}. For example, \citet{rawte2023exploring} explored how linguistic elements in prompts, particularly readability, formality, and concreteness, impact the occurrence of hallucinations. \citet{yao2023llm} illustrated that nonsense prompts, consisting of random tokens, can prompt LLMs to generate hallucinations, indicating that hallucinations might be viewed as another form of adversarial examples. \citet{manakul2023selfcheckgpt} detected hallucinations by evaluating the consistency of responses with BERTScore, QA-based metrics and n-gram metrics. In addition, several other studies rely on external knowledge for reference retrieval to detect hallucinations. \citet{chern2023factool} proposed a task and domain agnostic framework augmented by tools for detecting factual errors. \citet{yu2022generate} instructed a LLM to create contextual documents and then analyze the generated document to infer the ultimate answer. \citet{ren2023investigating} explored the perceptual capabilities of LLMs concerning the boundaries of factual knowledge through retrieval augmentation on open domain QA.

\paratitle{Hallucination Mitigation.} To mitigate the hallucinations, existing studies encompass researches across different stages in the development and utilization of LLMs. Specially, mitigation during pre-training is typically centered around dataset curating and cleaning. In this line, existing studies \citep{das2023diving,kamalloo2023hagrid,umapathi2023med} aim to construct a higher quality corpus for model pre-training by building datasets within specific  domains or cleaning existing datasets. After pre-training, fine-tuning approaches can be further employed for hallucination mitigation, such as the applications of SFT~\citep{wang2022self} and RLHF \citep{fernandes2023bridging}. In practical use, mitigation during generation is mainly focused on developing more effective decoding strategies~\citep{lee2022factuality,li2023inference,shi2023trusting,van2022mutual}, leveraging external knowledge~\citep{chern2023factool,varshney2023stitch} and designing more effective prompts~\citep{agrawal2023language,touvron2023llama}. Furthermore, mitigation during the post-processing stage can be  implemented by using LLM itself \citep{mundler2023self} or external knowledge~\citep{chen2023purr} as the fact-checking module to verify the generated text, so as to correct hallucinations. 

%% file: sec/conclusions.tex
\section{Conclusion and Limitation}

This paper presented a comprehensive empirical analysis about LLM hallucinations in the three aspects of detection, source and mitigation. 
We constructed the hallucination benchmark HaluEval 2.0 and developed an LLM-based automatic detection approach. Based on this benchmark, we further systematically investigated the possible sources for LLM hallucination in the stages of pre-training, SFT, RLHF, and inference, and also examined the effectiveness of a series of hallucination mitigation strategies, including RLHF, retrieval augmentation, self-reflexion, advanced decoding, and prompt improvement.  
As the major contribution, our benchmark can be reused for further research, and our work has led to a series of important empirical findings on the source and mitigation of LLM hallucination. 

Despite the great efforts that we have made, our analysis about pre-training stage and SFT is still limited, due to the lack of disclosed training details and the supporting computational sources. We will investigate into the two stages with more detailed analysis as future work. Furthermore, our experimental tests are not yet sufficient and we are also particularly interested in the working mechanism or nature of LLMs in generating the hallucinations. We will conduct more in-depth research work in the future. In addition, this paper mainly aims to provide empirical analysis on existing techniques to mitigate the LLM hallucinations, and there are no new hallucination mitigation strategies proposed. We will also consider developing improved mitigation strategies based on the findings of this work. Our hallucination detection approach is based on the GPT model (\ie GPT-4), which might inevitably lead to some minor errors. We will continuously improve our method in the future work.